\documentclass[conference]{IEEEtran}
\IEEEoverridecommandlockouts
\usepackage{cite}
\usepackage{amsmath,amssymb,amsfonts}
\usepackage{algorithmic}
\usepackage{graphicx}
\usepackage{textcomp}
\usepackage{xcolor}
\usepackage{float}
\usepackage{multirow}
\usepackage{tikz}
\usepackage{color, colortbl}
\usepackage{fancyhdr}
\definecolor{Yellow}{rgb}{1,1,0}

\def\BibTeX{{\rm B\kern-.05em{\sc i\kern-.025em b}\kern-.08em
    T\kern-.1667em\lower.7ex\hbox{E}\kern-.125emX}}

\newcommand{\omitit}[1]{}

\fancypagestyle{fancyfirst}
{
\fancyfoot[C]{
Distribution Statement A. Approved for public release: distribution is
unlimited. This work has been submitted to the IEEE for possible publication. Copyright may be
transferred without notice, after which this version may no longer be accessible.
}
}

\begin{document}

\title{Reward Shaping for Improved Learning in Real-time Strategy Game Play}

\author{\IEEEauthorblockN{John Kliem and Prithviraj Dasgupta}
\IEEEauthorblockA{
\textit{Distributed Intelligent Systems Section, Information Technology Division}\\
\textit{U. S. Naval Research Laboratory}\\
Washington, DC, USA\\
\{john.kliem, prithviraj.dasgupta\}@nrl.navy.mil}
}

\maketitle
\begin{abstract}
We investigate the effect of reward shaping in improving the performance of reinforcement learning in the context of the real-time strategy, capture-the-flag game. The game is characterized by sparse rewards that are associated with infrequently occurring events such as grabbing or capturing the flag, or tagging the opposing player. We show that appropriately designed reward shaping functions applied to different game events can significantly improve the player's performance and training times of the player's learning algorithm. We have validated our reward shaping functions within a simulated environment for playing a marine capture-the-flag game between two players. Our experimental results demonstrate that reward shaping can be used as an effective means to understand the importance of different sub-tasks during game-play towards winning the game, to encode a secondary objective functions such as energy efficiency into a player's game-playing behavior, and, to improve learning generalizable policies that can perform well against different skill levels of the opponent.
\end{abstract}

\begin{IEEEkeywords}
Reinforcement learning, reward shaping, RTS games, capture-the-flag
\end{IEEEkeywords}

\thispagestyle{fancyfirst}

\section{Introduction}
Deep reinforcement learning (RL) has been shown to be a very successful technique for enabling software agents to learn to play computer-based games including complex, real-time strategy (RTS) games~\cite{shao19}. Recent successes of deep RL include agents that can play single-player Atari games~\cite{mnih2013playing} and RTS games such as Starcraft II, DOTA 2, and Quake III Capture-the-flag (CTF)~\cite{vinyals2019grandmaster, dota2, JaderbergCDML19} with human champion level expertise. Learning to play RTS games  using deep RL presents several challenging problems such as sparse rewards, large state-action spaces, and correctly predicting the opponent's move so that an appropriate response move could be selected. Existing techniques to address these challenges in RTS games mainly utilize opponent modeling techniques. These include probabilistic tree search algorithms along with human user game-play information to determine suitable moves that guarantee higher expected rewards~\cite{silver2016mastering} and game-theoretic techniques such as self-play~\cite{silver2017mastering} and league play~\cite{vinyals2019grandmaster} to explicitly build opponent models and learn winning moves by training against them. While these techniques have shown promising results for board and RTS games they require considerable computing resources and learning times. For example, the Alpha-Star algorithm for playing Starcraft-II trained $900$ distinct agents on a cloud of $32$ tensor processing units (TPUs) per agent over a period of $44$ weeks and it required a dataset of $971,000$ game replays~\cite{vinyals2019grandmaster} in order to achieve better than human-expert performance.  It makes sense to investigate if less time and resource-intensive techniques could be used to learn to play RTS games without compromising performance.

To address this challenge, we investigate if reward functions, which are an inherent part of RL algorithms, could be used to encode the opponent's behavior and other game-related learning tasks to learn suitable game-playing strategies. Our main contribution in this paper focuses around a technique called reward shaping~\cite{Ng99} for successively adapting the reward function during training of the RL agent guided by the player's observation of the opponent and the win/loss states within a game. To the best of our knowledge, our technique is one of the first detailed studies of reward shaping to improve learning performance in RTS games. We evaluate our reward shaping technique where a player learns to play a 1-v-1, marine Capture-the-Flag (CTF) game called Aquaticus CTF, against two skill levels of opponents. We have validated our results using the MOOS-IvP simulation environment for marine robots. Our results show that using reward shaping, a player can successfully learn to play the game faster, achieve higher scores, and adapt its play suitably to different levels of opponents, than without reward shaping. From our results, we also make three novel observations about the impact of reward shaping: (1) reward shaping can enable the learning agent to identify which sub-tasks are more important to learn towards the overall task of winning the game, and appropriately focus the reward shaping function towards those sub-tasks, (2) reward shaping can be used to provide a suitable means to encode additional objectives for a player that are not inherent in the game rules, for example, moving in an energy efficient manner while playing the game, and, (3) reward shaping enables a player to learn more generalizble policies. The rest of the paper is structured as follows: in the next section we discuss relevant literature on reward shaping for RTS games. Section~\ref{sec:aquaticus_ctf} describes the Aquaticus CTF game setting and rules, and, the formal framework for the game. In Sections~\ref{sec:reward_shaping} and~\ref{sec:expts} we describe the different reward shaping functions relevant to the Aquaticus CTF game and our experimental validation of the effects of the reward shaping functions on the player's score and training times, and, finally, we conclude.

\section{Related Work}
Reward shaping has been proposed as a lightweight yet effective means to focus an RL agent's exploration of the environment during training towards regions that enable it to reach its goal quickly. The main idea in reward shaping is to update the nominal or sparse reward at a state-action pair as a function of the suitability of the state-action pair towards reaching the goal~\cite{grzes17rewardshaping}. In one of the earliest works on reward shaping~\cite{Ng99}, authors proposed this function as the difference of a potential function between the next and current states, which was approximated using the negative Manhattan distance to the goal between the two states. Subsequent researchers have extended this idea by proposing variations to the potential function using future as well as past states~\cite{Wiewiora03}, using distance to nearest states given in expert demonstrations~\cite{Brys15}, using a bi-level optimization problem to simultaneously determine optimal values for the potential function and policy parameter (e.g., weights of a policy network) of the learning problem~\cite{hu2020learning}, and, using a technique called exploration guided reward shaping~\cite{devidze2022exploration} where the reward of a state-action pair is dynamically updated based on the gradient of an objective function along with a bonus based on state visit counts. Most of these techniques have been proposed for single agent tasks in a non-game playing environment. Our work builds on these techniques by applying them to a competitive, 2-player game. Related to our work, Jaderberg {\em et al.}~\cite{JaderbergCDML19} showed that agents could learn to play CTF while using reward shaping to adaptively learn to respond to abrupt changes in the game rules or environment. They proposed reward shaping functions around $12$ consequential game-events such as grabbing, capturing and retrieving the flag or tagging an opponent. Their work's main focus was on learning winning strategies while we focus on the specific impacts of reward shaping in the CTF game. Another minor difference is that our RL algorithm uses physically observable features of players like locations and headings that can be readily mapped to physical robots and robots' on-board sensor data, in contrast to using derived or private information like player scores and pixel maps of the game for learning.
We have used reward shaping as a means to model opponent behavior. The topic of opponent modeling for learning to play computer games has also been researched actively in recent years. Early approaches include self-play in repeated games and its deep learning-based counterpart called fictitious neural self-play~\cite{heinrich16nfsp}. More recently, He {\em et al.}~\cite{he2016opponent} proposed the Deep Reinforcement Opponent Network (DRON) algorithm that uses opponent behavior traces to update the player's DQN parameters directly or via a mixture of experts. In~\cite{FoersterCAWAM18,willi2022cola}, authors proposed the learning with opponent learning awareness (LOLA) algorithm that updates the gradient of a player's policy network  while including both its own and its opponent's value functions and policy gradient updates. Opponent modeling as an auxiliary task was proposed in~\cite{Hernandez-LealK19} where players utilized a central policy network with parameter sharing but different outputs for each player. The technique was mainly validated for smaller, grid-like environments of games. Also, in contrast to our work, these approaches directly update the player's policy network's gradients or parameters to account for the opponent's actions.
\section{Aquaticus Capture-the-Flag Game}
\label{sec:aquaticus_ctf}
\begin{figure} [t]
    \centering
    \includegraphics[width=3.4in]{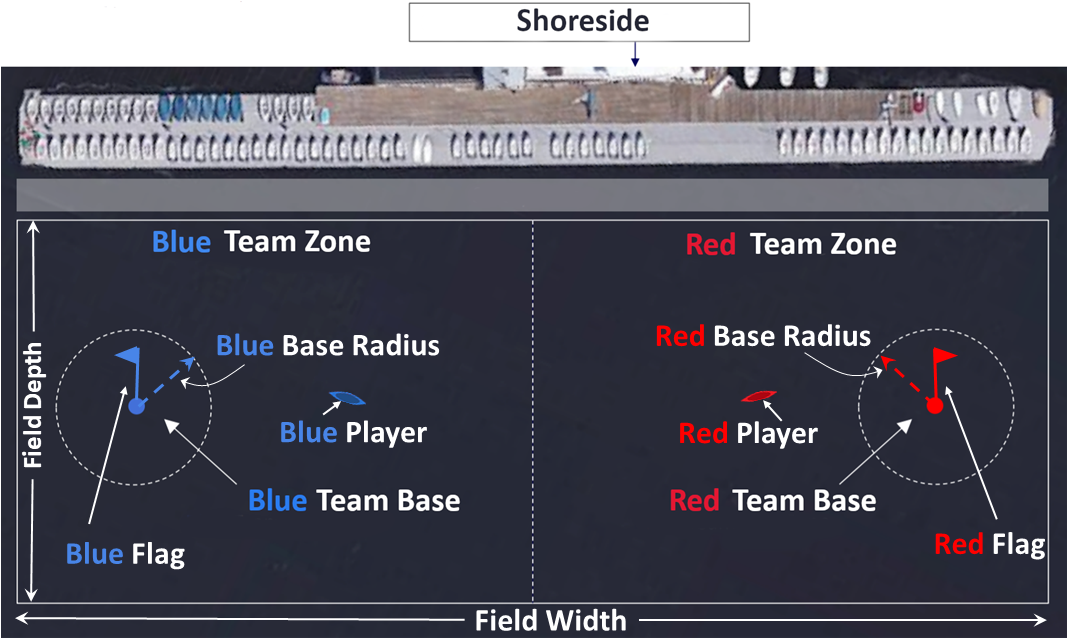}
    \caption{Aquaticus CTF Game Field.}
    \label{fig:aquaticus_ctf_v2}
\end{figure}

Capture the Flag (CTF) is a classic game that is played between two opposing teams. We consider a marine, $1$-v-$1$ version of the CTF game called Aqaticus CTF between an attacker (red) and a defender (blue). Figure~\ref{fig:aquaticus_ctf_v2} shows an image of the Aquaticus CTF playing field. The game is played in a rectangular, obstacle-free playing field that is divided into two halves, which we will refer to as a player's zone. Each player has a base within its zone that contains the player's flag. The objective of the attacker is to pick the flag from the defender's base (called flag grab) and return with it back to its own base (called flag capture), without being tagged by the defender during this process. The defender's objective is to tag the attacker while the attacker is inside the defender's zone. A player gets tagged by its opponent when it gets within the tag range of its opponent inside the opponent's zone. Tagging a player forces the tagged player to return to its base to get un-tagged. Similarly, if a player goes out-of-bounds (OOB) of the playing area during the game{\footnote{A player could go OOB either deliberately (e.g., to avoid collision with another player) or inadvertently (e.g., due to limitations of its vehicle's dynamic capabilities like turn radius, inaccuracies of its on-board localization system, etc.).}}, it has to revert to its base before starting to play again. A player going OOB or getting tagged does not pause the game and other players inside the playing area continue playing the game. The game events and corresponding scoring rules for the Aquaticus CTF game are given in Table~\ref{tab:ctf_game_outcomes}. There is a game arbiter called shoreside that has full observation of the game, registers the different events as they occur in the game, maintains players' scores, and facilitates communication of the game states across players.

\begin{table}[htb!]
    \centering
    \caption{Events and points in the Aquaticus CTF game.}
    \begin{tabular}{|c|c|c|c|}
        \hline
        {\bf Event Name} & {\bf Event} & {\bf Attacker} & {\bf Defender}\\
                        & {\bf Description} & {\bf Points} & {\bf Points}\\
        \hline
        \hline
        Tag & Defender tags attacker before & $-1$ & $+2$\\
            & grab or attacker goes OOB     &   &  \\
        \hline
        Retrieval Tag & Defender tags attacker & $-2$ & $+1$\\
                        &   after grab          &       & \\
        \hline
        Flag grab & Attacker picks flag & $+1$ & $-1$ \\
                  & from defender's base & & \\
        \hline
        Flag capture & Attacker returns to its & $+2$ & $-2$\\
                     &  base with grabbed flag     &          & \\
        \hline
        Defender tag & Defender tagged by attacker   & $+2$ & $-2$\\
                        & or defender goes OOB   &      & \\
         \hline
             \end{tabular}

    \label{tab:ctf_game_outcomes}
\end{table}
The game goes on for a pre-determined, fixed duration and comprises of one or more rounds. A round starts with the attacker and defender starting at their base regions. A round ends when one of the following events occurs:
\begin{enumerate}
    \item The attacker does a successful flag capture.
    \item The attacker gets tagged by the defender while trying to do a flag grab or flag capture.
    \item The game duration ends before either of the aforementioned events occur
\end{enumerate}
Note that by this definition, the duration of rounds, and consequently the number of rounds in a game is not fixed. 

{\bf Player Learning Tasks.} A player that is learning to play the game has multiple learning tasks. Each player's primary task is to learn to defeat its opponent: the defender learns how to move so that it can tag the attacker quickly before the attacker grabs or captures the defender's flag; the attacker learns how to move so that it can grab and subsequently capture the flag without getting tagged by the defender. The players are also not directly provided with the location of the boundaries of the field. A secondary learning task for each player is to be able to discover these boundaries so that it does not accrue negative points due to going out-of-bounds (as per Table~\ref{tab:ctf_game_outcomes}), and learn to remain within the playing area during the game. Finally, we also explore additional learning tasks for players such as how to determine moves that reduce energy expenditure while playing.

\subsection{Formal Game Representation}
\begin{table}
\centering
\caption{Observation and game parameters in the Aquaticus CTF game.}
\begin{tabular}{|c|l|}
\hline
{\bf Notation} & {\bf Description}\\
\hline
$\mathbf{X}^{field}$ & Rectangular playing field of size \\
					& $ D \times W$\\
\hline
$\mathbf{L}^{bounds} = \{L^{upper},$ & Upper, lower, left and right boundaries \\
$L^{lower}, L^{left}, L^{right}\}$ &of playing field \\
\hline
$Z^{att}, Z^{def}$ & Attacker and defender zones (halves)\\
\hline
$\mathbf{x^{att}}, \theta^{att}$ & Attacker location and heading\\
\hline
$\mathbf{x^{def}}, \theta^{def}$ & Defender location and heading\\
\hline
$x^{aflg}, x^{dflg}$ & Location of attacker and defender flags\\
\hline
$D^{base}$ & Base radius around player's flag \\			
\hline
$D^{tag}$ & Minimum distance between player and \\
			& opponent locations for tagging\\
\hline
$D^{grab}, D^{cap}$ & Min. dist. between player and flag \\
			& or base locations for a flag grab/capture\\
\hline
\hline
$D^{warn}$ 	& Minimum distance between player and \\
			& boundary or opponent for warning\\
\hline
$D^{threat}$ & Minimum distance between player and \\
			& boundary or opponent for threat\\
\hline
\end{tabular}
\label{table:parameters}
\end{table}

The environment of the Aquaticus CTF game is represented by a rectangular, obstacle-free playing area of width $W$ and depth $D$ bounded by a set of four straight lines, $\mathbf{L}^{bounds} = \{L^{lower}, L^{upper}, L^{left}, L^{right}\}$. Attacker and defender zones and bases are denoted by $Z^{att}$ and $Z^{def}$, and, $B^{att} \subset Z^{att}$ and $B^{def} \subset Z^{def}$, respectively. The attacker's and defender's base regions are denoted by $\mathbf{X}^{abase}$ and $\mathbf{X}^{dbase}$ and flag locations within respective bases are denoted as $\mathbf{x}^{aflg}$ and $\mathbf{x}^{dflg}$. The different parameters used in the Aquaticus CTF game are shown in Table~\ref{table:parameters}. In the rest of the paper, for the sake of brevity, we have given notations and definitions for the defender, assuming that the corresponding notations and definitions for the attacker are commensurately defined.

The defender's state set is given by $S^{def} = \{(\mathbf{x^{def}}, \theta^{def})\}$, where $\mathbf{x^{def}} = (x^{def}, y^{def}) \in \mathbb{Z}^2$ specifies its location in the playing field. $\theta^{def} \in \Theta = \{\theta_1, \theta_2, \ldots, \theta_K\}$ is a discretization of the angular bearing  $[-\pi, \pi]$ into $K$ disjoint segments. The action set of the defender is specified as $A^{def} = \{(\nu, \theta)\}$, where $\nu \in \{0 \cup \Re^+\}$ and $\theta \in \Theta$ are the defender's velocity and heading respectively. The attacker's state and action sets are defined commensurately and denoted by $S^{att}$ and $A^{att}$. The state space of the game is represented as the joint defender and attacker state sets, $S = S^{def} \times S^{att}$. The action space of the game is given by the joint actions of the defender and attacker, $A = A^{def} \times A^{att}$. The forward dynamics model or transition function ${\mathbb T}: S \times A \times S \rightarrow [0,1]$ gives a probability distribution for the next game state given the current game state and current actions taken by the players. Finally, for legibility, we define a subset of states corresponding to the point-scoring game events in Table~\ref{tab:ctf_game_outcomes}, $S^{EV} = \{S^{tag}, S^{ret}, S^{oob}, S^{grb}, S^{cap}\}$. The formal definition of these sets in given in the supplementary material.


\subsection{Reinforcement Learning for Aquaticus CTF}
We model the Aquaticus CTF game as a Markov game (or stochastic game)~\cite{shapley1953stochastic} represented as $(S, A, \mathbb T, R, \gamma)$, where $S, A, \mathbb T$ are the state and action spaces, and transition function of the game as defined above. $\gamma \in [0, 1]$ is a discount factor and $R: S \times A \rightarrow \Re$ is the players' joint reward function. $R = (R^{att}, R^{def})$ where $R^{def}$ and $R^{att}$ are the defender's and attacker's reward functions respectively. Each player's objective in the game is to determine an action sequence for itself that maximizes its rewards so that it ends up in states that result in its getting more points than its opponent and winning the game. Because of the competitive nature of the game, as shown by the players' points structure in Table~\ref{tab:ctf_game_outcomes}, higher rewards for one player at a certain state correspond to lower rewards for its opponent at the same state and vice-versa. This makes each player's objective contrary to its opponent's objective. To calculate a winning action sequence, each player uses an RL algorithm. A player's RL algorithm takes the stochastic game parameters as input and outputs a policy that prescribes the action for the player at each state in the game. We denote $\pi^{def}$ and $\pi^{att}$ as the defender and attacker policies, respectively. $\pi^{def}: S \rightarrow \Delta(A^{def})$, where $\Delta(A^{def})$ is a probability distribution over $A^{def}$. $\pi^{att}$ is defined correspondingly.


A trajectory in the game is given by a sequence of state-action-reward tuples over successive time-steps, $\tau = ((s_0, a_0, r_0), (s_1, a_1, r_1), ..., (s_{T-1}, a_{T-1}, r_{T-1}), (s_T, \{\varnothing\} , r_T))$, where $s_t \in S$, $a_t =  (a^{att}_t, a^{def}_t) \in A$, $s_{t+1} = \mathbb T(s_t, a_t)$, and $r_t \in R$. Here $s_t$, $a_t$ and $r_t$ represent the game state, the joint action taken by the players and the joint reward received by them in the $t$-th time-step. $(a^{att}_t, a^{def}_t) = (\pi^{att}(s_t), \pi^{def}(s_t))$. $T$ denotes the trajectory length measured in number of time-steps. A trajectory starts when either the player or its opponent starts from its base to attack or defend, and terminates when any one of the events in Table~\ref{tab:ctf_game_outcomes}, except 'Flag grab' occurs. A game consists of a sequence of trajectories $(\tau_1, \tau_2, \tau_3... )$. Note that by the above definition of trajectory, the length of a trajectory can vary depending on when a trajectory ending event happens. Also, since a game goes on for a fixed duration, the number of trajectories in a game could vary depending on the length of each trajectory in it. For legibility, we denote the defender's state-action-reward components in trajectory $\tau$ as the defender trajectory, $\tau^{def} = (s^{def}_t, a^{def}_t, r^{def}_t): t = 0.. T$, and, correspondingly the attacker trajectory as $\tau^{att}$. 

The defender's return from defender trajectory $\tau^{def}$ is $G_{\tau^{def}} = \sum_{t=0}^T r^{def}_t$. The attacker's return is correspondingly $G_{\tau^{att}} = \sum_{t=0}^T r^{att}_t$. The value of a defender policy $\pi^{def}$ when the attacker uses attacker policy $\hat{\pi}^{att}$ is given by $V^{\pi^{def}|\hat{\pi}^{att}} = {\mathbb E}_{\tau^{def} \sim \pi^{def}} [G_{\tau^{def}}]$. Note that although attacker actions are not explicitly included in the r.h.s. of the above equation, it indirectly influences $G_{\tau^{def}}$ as $(s^{def}_{t+1}, s^{att}_{t+1}) = \arg \max_{s' \in S} \mathbb T(s_t, (\pi^{def}, \hat{\pi}^{att}), s')$. Let $\Pi^{def}$ denote the set of all possible defender policies. The optimal defender policy in response to attacker policy $\hat{\pi}^{att}$ is given by $\pi^{def^*}|\hat{\pi}^{att} = \arg \max_{\pi^{def} \in \Pi^{def}} V^{\pi^{def}|\hat{\pi}^{att}}$. 

To evaluate the performance of a defender trajectory $\tau^{def} \sim \pi^{def}$, we define a trajectory score $J(\tau^{def})$. We denote the number of tag events in $\tau^{def}$ as $N^{tag} = \#(\{s^{def}_t: s^{def}_t \in tau^{def} \wedge s^{def}_t \in S^{tag}\}$, where the function $\#()$ returns a count of the elements passed to it. Similarly, we define counters for other game events in the trajectory, namely, retrieval tags, out-of-bound events, grab and capture events as $N^{ret},  N^{oob}, N^{grb}$ and $N^{cap}$ respectively. The trajectory score is then given by:
\begin{eqnarray}
\label{eqn:score_def}
J(\tau^{def}) & = & c(S^{tag}) N^{tag} + c(S^{ret})N^{ret} + c(S^{oob}) N^{oob}\nonumber \\
              &  & + c(S^{grb}) N^{grb} + c(S^{cap})\,N^{cap}, 
\end{eqnarray}
\noindent where $c()$ gives the points for the events corresponding to the states from Table~\ref{tab:ctf_game_outcomes}. The score of defender policy $\pi^{def}$ is given by $J(\pi^{def}) = \mathbb E_{\tau^{def} \sim \pi^{def}} [J(\tau^{def})]$.

\subsection{Sparse Reward Functions}
A player's reward function is a central component of its RL algorithm as the policy output by the RL algorithm is calculated by maximizing the player's expected returns. The Aquaticus CTF game, like most RTS games, is characterized by a sparse reward structure where the players receive non-zero rewards only at states corresponding to the game events in Table~\ref{tab:ctf_game_outcomes} and zero at all other states. To model this sparse reward structure, we define a player's reward function as consisting of two components - an internal and an external, as given below:
\begin{gather}
R^{def} = R^{def}_{int}(s, a) + R^{def}_{ext} (s) \nonumber \\
R^{def}_{int}(s, a) = 0, \quad \quad \forall s \in S \nonumber \\
R^{def}_{ext}(s) = \begin{cases}
    C_{ext} c(s), & \text{if } s \in S^{EV} \\
    0, & \text{otherwise}
\end{cases} 
\label{eqn:reward_def_ext_int}
\end{gather}
where  $R^{def}_{int}$ is an intrinsic reward that is internal to each player, and $R^{def}_{ext}$ is an external component determined by the game rules. If the game state $s \in S^{EV}$, corresponding to one of the events in Table~\ref{tab:ctf_game_outcomes}, $R^{def}_{ext}(s)$ equals to the points for that event in the table, $c(s)$, scaled up by a constant scaling factor $C_{ext}$, otherwise it is $0$. $R^{att}$ is correspondingly defined.

\section{Reward Shaping Around Target and Avoid States}
\label{sec:reward_shaping}
\begin{figure}
    \centering
    \begin{tabular}{cc}
    \hspace{-0.1in}
    \includegraphics[height=1.5in, width=1.7in]{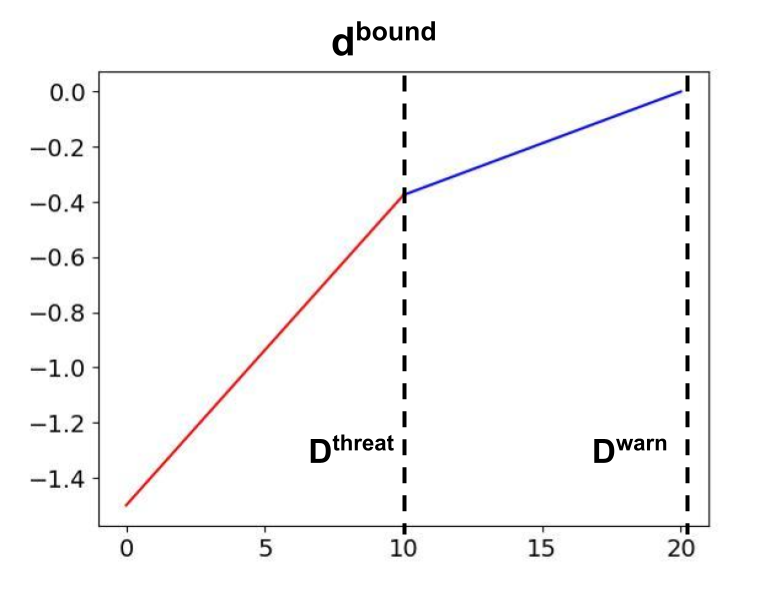} &
    \hspace{-0.15in}
    \includegraphics[height=1.5in, width=1.7in]{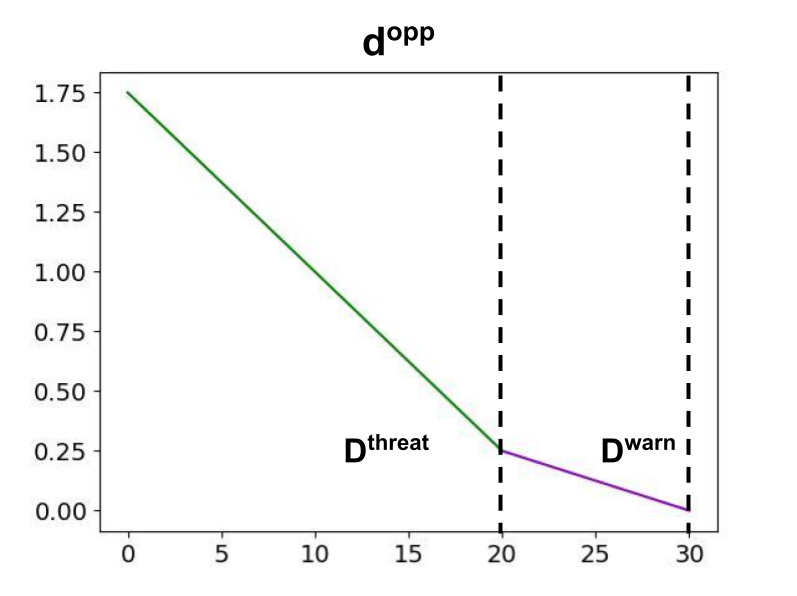} \\
    {\small (a)} & {\small (b)}\\
    \end{tabular}
    \caption{Example graphs for (a) boundary and (b) tag reward shaping functions.}
    \label{fig:rewardshapingppofig}
\end{figure}

Reward shaping augments the intrinsic component of a player's sparse reward function at a state-action pair with a value that represents the suitability of the state-action towards reaching the goal. Formally, the shaped reward, $\hat{R}(\cdot)$ is given by: $\hat{R}(s_t, a_t, s_{t+1}) = R(s_t, a_t, s_{t+1}) + \rho(s_t, a_t, s_{t+1})$ where $R(\cdot)$ is the sparse reward, and $\rho(\cdot)$ is a reward shaping function. Following~\cite{Ng99}, we have used $\rho(s_t, a_t, s_{t+1}) = \gamma\phi(s_{t+1}) - \phi(s_t)$, where $\phi(s)$ gives the potential, or informally, the suitability of state $s$ towards reaching a target state, and $\gamma$ is a discount factor. We consider three different reward shaping functions{\footnote{We have used piece-wise linear functions as approximations of polynomial or exponential functions to speed-up calculations as the reward shaping function is computed and added to the rewards at each iteration of a training step.}} corresponding to principal tasks to be learned by the players, as described below:

{\bf Boundary Reward Shaping.} Learning the locations of the boundaries of the playing field is one of the most difficult tasks for players as the sparse reward function has an abrupt drop or `cliff' at the boundary from zero to $C_{ext}*c(s^{oob})$. In the worst case, to learn the location of all the boundaries, the player would need to traverse from the inside to the outside of the field across every point on each of the boundary lines during training  A training round also terminates as soon as the player that is learning leaves the field, and so, the learning player would need to have an enormous number of  training rounds to completely learn the boundary. To ameliorate the significant training times we associate a gradient with the reward function near the boundaries, representing a slow drop off instead of a cliff in the reward function at the boundaries. The boundary reward shaping function implements a negative or downward gradient towards the boundaries from inside the field and is given by:

\begin{equation}
     \rho^{bnd}= 
\begin{cases}
    -\kappa_1 + \kappa_2 \, d^{bound},& \text{if } {D^{threat} \leq d^{bound} < D^{warn}}\\
    -\kappa_3 + \kappa_4 \, d^{bound},& \text{if } {0 \leq d^{bound} < D^{threat}} \\
     0, & \text{otherwise}
\end{cases}
\label{eqn:reward_shaping_bound}
\end{equation}
    
\noindent where $d^{bound}$ is the distance of the player to its closest boundary and $\kappa_1, \kappa_2, \kappa_3, \kappa_4 \in \Re^+ $ are positive constants. Figure~\ref{fig:rewardshapingppofig}(a) illustrates an example of the boundary reward shaping function.

{\bf Tag Reward Shaping.} Tagging can be considered as an attractive potential towards the opponent. With sparse reward for tagging, the learning player encounters an abrupt increase in reward from zero to $C_{ext}*c(s^{tag})$ when it gets at a tag state. Relying only on value updates gradually guides the defender by building an attractive potential towards the tag states, but again requires training rounds. With a reward shaping function that provides an attractive gradient towards tag states, we can accelerate the value updates so that the defender can recognize tag states when it is further away from them and get attracted to them quickly, resulting in improved tagging performance by the defender. The tagging reward shaping function we have used implements a positive or upward gradient towards tag states, as given below:

\begin{equation}
    \rho^{tag}= 
\begin{cases}
    \omega_1 - \omega_2\, d^{opp},& \text{if } {D^{threat} \leq d^{opp} < D^{warn}}\\
    \omega_3 - \omega_4\, d^{opp},& \text{if } {D^{tag} \leq d^{opp} < D^{threat}}\\
    0, & \text{otherwise}
\end{cases}
\label{eqn:reward_shaping_tag}
\end{equation}

\noindent where $d^{opp}$ is the distance between a player and its opponent and $\omega_1, \omega_2, \omega_3, \omega_4 \in \Re^+ $ are positive constants. Figure~\ref{fig:rewardshapingppofig}(b) illustrates an example of the tag reward shaping function.

{\bf Energy Reward Shaping.} A latent task of the players is to reduce the energy they expend for moving around while playing the game. In computer games, moving is usually free, but if we consider games with physical agents like marine robots, frugal energy expenditure enables the robotic players to persist longer in the game. Players could reduce their energy expenditure if they could receive a reward signal that motivates them to follow shorter paths towards their destination states in the game, or to remain stopped at strategic locations and move towards their goal only when necessary. Towards this objective, our energy reward shaping function gives a larger positive reward for low energy actions like stopping, a slightly positive reward for medium energy actions like continuing the action from the previous time-step (e.g., moving at constant speed or remaining stopped), and a negative reward for higher energy actions like changing speed or making turns, as given below:
\begin{equation}
\rho^{enrg}= 
\begin{cases}
    \mu_1,& \text{if } \nu =0\quad\text{(player stopped) and }  a_{t-1} = a_t\\
    \mu_2,& \text{if } a_{t-1} = a_t\\
    -\mu_3,& \text{if } a_{t-1} \neq a_t\\
\end{cases}
\label{eqn:reward_shaping_enrg}
\end{equation}    

\noindent where $\mu_1, \mu_2, \mu_3  \in \Re^+$ are positive constants. 

{\bf Reward Shaping for Learning Generalizable Policies.} In RTS games, the skill level of players on the opposing side is not known {\em a priori} and an intelligent player should be able to adjust its playing strategy to win against a diverse set of opponent skill levels. To achieve this, we investigate if reward shaping can used to improve learning of policies that generalize well against different skill levels of opponents. We investigate two training methods with and without reward shaping: interleaved training and curriculum learning. In interleaved training, we train a player against a randomly selected opponent skill level in each training episode. In curriculum learning, a player is trained sequentially against each opponent skill level. The trained policy against one opponent skill level is used as the initial policy for training against the next opponent skill level.

\section{Experimental Results}
\label{sec:expts}
\begin{figure}[t]
    \centering
    \includegraphics[width=3.5in]{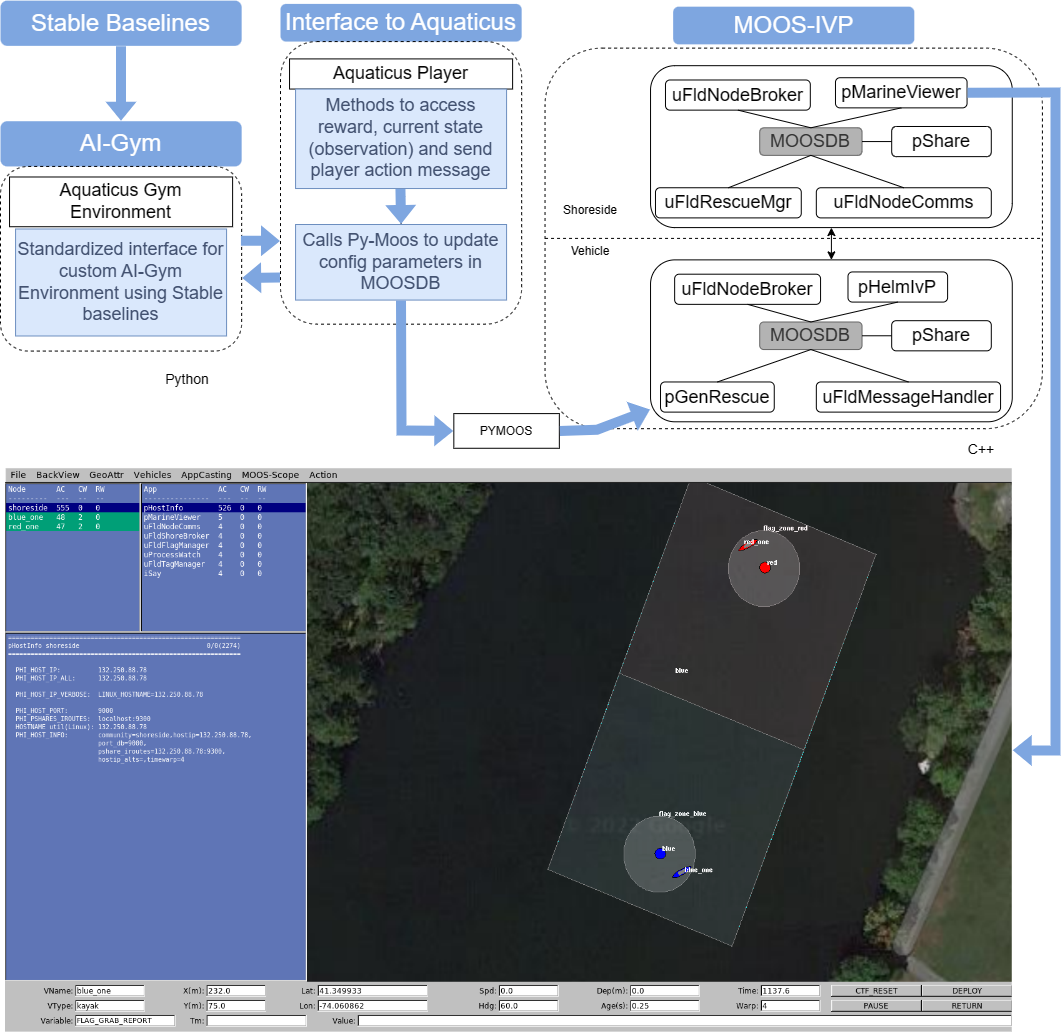}
    \caption{Software architecture of the Aquaticus Gym and Moos-IvP (top) used in our experiments and a screenshot of the Aquaticus CTF game (bottom).}
    \label{fig:amoos}
\end{figure}

{\bf Game environment.} For our experiments we have used the Aquaticus-Gym CTF game environment that uses MOOS-IvP~\cite{benjamin2013autonomy} as the robotic simulation platform. MOOS-IvP simulates the dynamics of robotic platforms within a marine movement and provides a message-based communication framework between the robots. The Aquaticus-Gym CTF layer handles the CTF game rules along with an interface for RL algorithms. It receives the current game state information from MOOS-IvP and returns an action (speed, heading) for each player using the learned policy. Figure~\ref{fig:amoos} shows a schematic of the interaction between the Aquaticus-Gym CTF and MOOS-IvP components of the software.The parameters in the game are field width, $W=160\,$m, field depth, $D=80\,$m. This makes each player's zone's size $80 \times 80\,$m$^2$. The base radius ($D^{base})$, tag range ($D^{tag}$) and grab range ($D^{grab}$) are $10\,$m each. Game durations measured in clock time are $40$ minutes during training and $60$ minutes during evaluation; simulations were run at $4\times$ speed-up. The initial parameters used in the reward shaping equations, Equations~\ref{eqn:reward_shaping_bound}-~\ref{eqn:reward_shaping_enrg} are $\kappa_1=-0.1875$, $\kappa_2=0.0.028125$, $\kappa_3=-0.375$, $\kappa_4=0.0125$, $\omega_1=0.1875$, $\omega_2=0.028125$, $\omega_3=-0.375$, $\omega_4=0.075$, $\mu_1=0.5$, $\mu_2=0.4$, and, $\mu_3=-0.5$.

We experimented with two deep RL algorithms Proximal Policy Optimization (PPO)~\cite{schulman2017proximal} and Deep Q-network (DQN)~\cite{mnih2015human} for training a defender agent with different reward shaping functions. We chose these two algorithm as PPO has been recently reported to improve learning in RTS games without requiring excessive hyper-parameter tuning or domain-specific algorithm modifications~\cite{yu2022surprising} while DQN has been one of the most successful algorithms for learning to play Atari games!\cite{mnih2013playing}. We have used the implementation of these algorithms available in the Stable-Baselines 3 library~\cite{stable-baselines}. In this section, we discuss the results with PPO as it yielded faster convergence and better game-playing performance; results with DQN are given in supplementary material. Our PPO policy network uses a multi-layer perceptron with $2$ layers of $64$ perceptrons each. The training hyper-parameters of the PPO algorithm are: discount factor, $\gamma = 0.99$, learning rate, $\alpha = 0.0005$, and buffer size$=5000$. All our experiments were run on a server with $200$GB RAM, four $2.7$GHz $12$-core Intel Xeon CPUs each with $19$ MB cache, two Quadro P$5000$ GPUs each with $16$ GB RAM, with Ubuntu $20.10$ as the operating system. Each PPO training was done for a minimum of $600$K time-steps, corresponding to roughly $16$ hours of clock time for each training.

{\bf Experimental Setup.} To clearly discern the effect of reward shaping, we have evaluated our reward shaping techniques while training only one side, a defender agent, against two levels of attackers, easy and hard. Both attackers use a rule-based play strategy. The easy attacker (Att-E) follows a fixed trajectory going back and forth between its own base and the defender's base and flag. It is agnostic of the defender's presence or location and does not avoid getting tagged. The hard attacker (Att-H) uses a potential fields-based strategy to actively avoid getting tagged by the defender while finding a trajectory to grab the flag or get back to its zone and base. 
\begin{figure*}
    \centering
    \begin{tabular}{ccc}
    \includegraphics[width=2.5in]{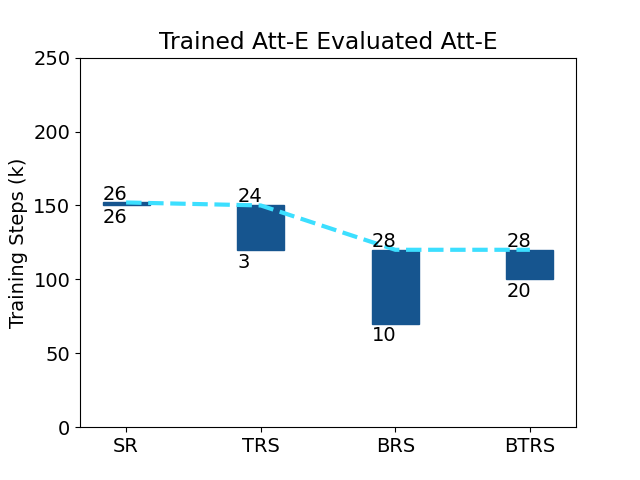} &
    \hspace{-0.3in}
    \includegraphics[width=2.5in]{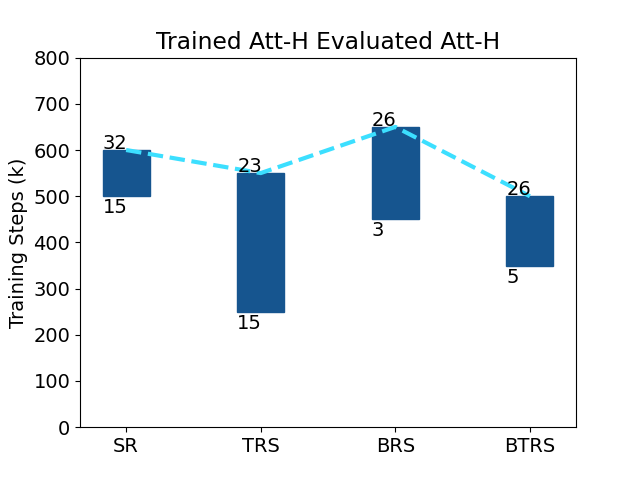} &
    \hspace{-0.3in}
    \includegraphics[width=2.5in]{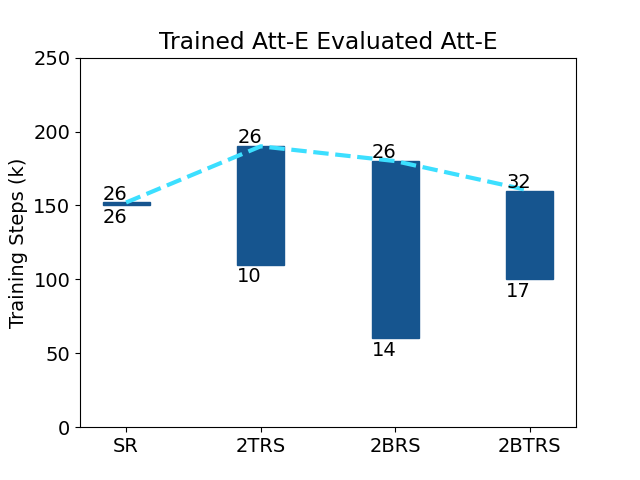} \\
    (a) & (b) & (c) \\
    \includegraphics[width=2.5in]{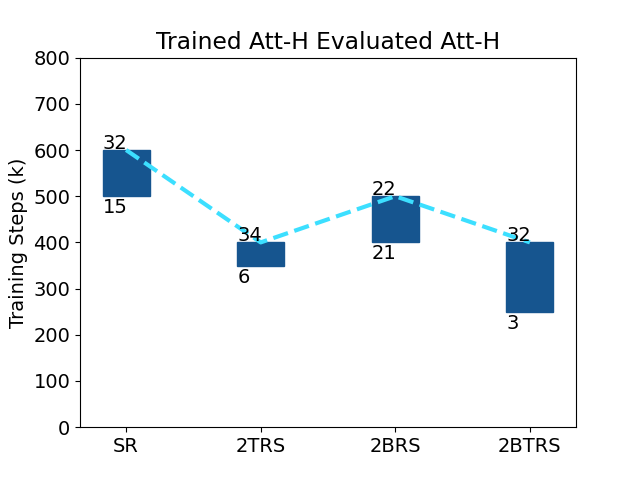} &
    \hspace{-0.4in}
    \includegraphics[width=2.3in]{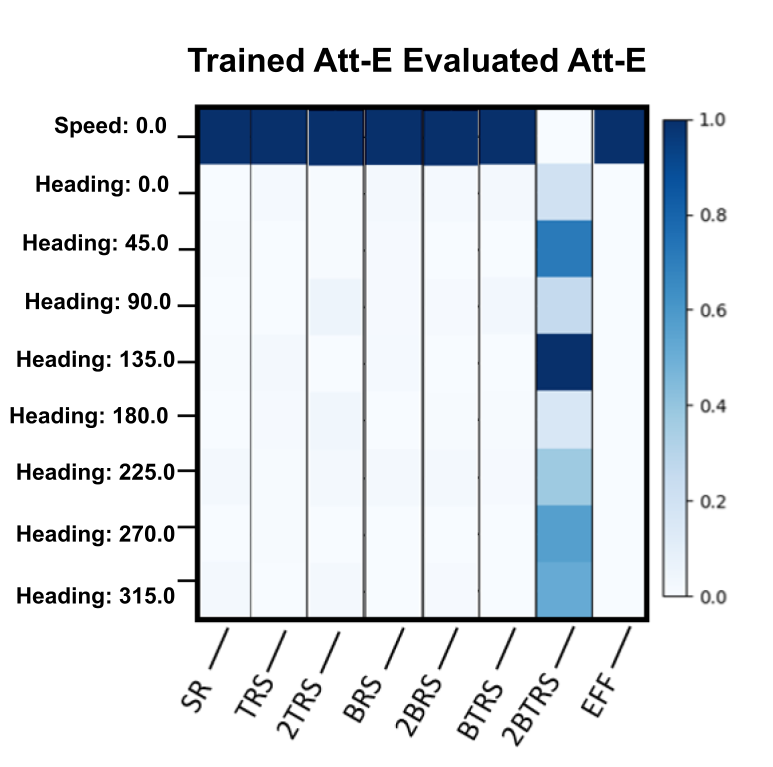} &
    \hspace{-0.2in}
    \includegraphics[width=2.3in]{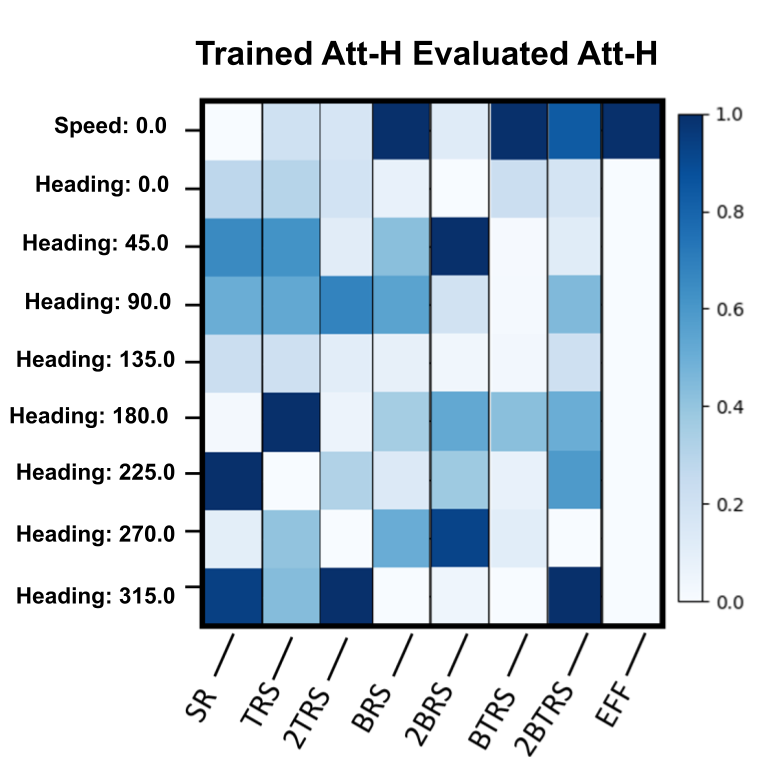} \\

    (d) & (e) & (f) \\
    \end{tabular}
    \caption{(a) - (d) Training times (y-axis) and scores of a defender at the start and end of score convergence using {\tt SR} and reward shaping functions  {\tt TRS}, {\tt BRS} and {\tt BTRS} that is trained and evaluted against: (a) Att-E attacker, with $1X$ gradients of reward functions, (b) Att-E attacker, with $2X$ gradients of reward functions, (c) Att-H attacker, with $1X$ gradients of reward functions, (d) Att-H attacker, with $2X$ gradients of reward functions. (e) and (f) action heat maps of the defender trained against Att-E and Att-H respectively using {\tt SR} and reward shaping functions  {\tt TRS}, {\tt BRS} and {\tt BTRS} with $1X$ and $2X$ gradients.}
    \label{fig:ppo_training_score}
\end{figure*}

\subsection{Experimental Evaluation.} 
For each of our experiments, we used the sparse reward ({\tt SR)} as the baseline, and evaluated the defender agent's score while using the tag ({\tt TRS}), boundary ({\tt BRS}), and a combination of boundary and tag ({\tt BTRS}) reward shaping functions. Learning defender agents were evaluated at intervals of $10,000$ training steps against Att-E and at intervals of $50,000$ training steps against Att-H. To understand the effects of the reward shaping techniques, we have evaluated four hypotheses, as given below:\\
{\bf Hypothesis 1.} Reward shaping improves the performance of the defending player in terms of reward convergence times and improved scores, as compared to using sparse rewards only.\\
{\bf Hypothesis 2.} When an agent has to learn multiple sub-tasks as part of its overall task, reward shaping for the more difficult tasks results in higher improvements in the agent's performance than reward shaping for the less difficult tasks.\\
{\bf Hypothesis 3.} Reward shaping can be used to encode additional objective functions like energy conservation while ensuring consistently positive scoring policies when playing the game. \\
{\bf Hypothesis 4.} Adjusting gradients of the reward shaping function can guarantee faster convergence to higher rewards. \\
{\bf Hypothesis 5.} Reward shaping improves learning generalizable policies.

{\bf Hypothesis 1.} {\em Reward shaping improves the performance of the defending player in terms of reward convergence times without degrading its performance, as compared to using sparse rewards only.}  To validate Hypothesis 1, in our first set of experiments, we trained a defender agent against Att-E and Att-H using the different rewards functions within the PPO algorithm. The results are shown in Figure~\ref{fig:ppo_training_score} (a)-(b). Each vertical bar in the graphs represents the number of time steps between the time the defender's reward starts to converge and the time when the reward converges for each reward shaping function. The numbers at the bottom and top of each bar give the defender's scores at those times respectively. Against, Att-E, the trend lines of the final reward convergence times shows a downward gradient indicating that the reward shaping functions improve the learning times as compared to the {\tt SR} baseline. The final score achieved by the defender also remains stable or decreases nominally against Att-E indicating that the improved learning times do not degrade the defender's performance. However, against the harder Att-H, while the learning times again exhibit a downward trend line, the final score of the defender decreases more aggressively by about $20\%$ for different the reward shaping functions. To investigate the degrading scores of the defender against Att-H, we adjusted the gradients of the reward shaping functions in Equations~\ref{eqn:reward_shaping_bound}-\ref{eqn:reward_shaping_tag} to lower values ($0.5\times$) and higher values ($2\times$). The corresponding results with $2\times$ gradients are in Figure~\ref{fig:ppo_training_score} (d) and show that the defenders reward converge times improve without degrading the scores. However, with $2\times$ gradients, the defender's learning times and performance against Att-E get degraded, as seen in Figure~\ref{fig:ppo_training_score} (c). This indicates that when the opponent's behavior changes, the defender's reward shaping function's gradient needs to be carefully adjusted to continue to yield good performance in the game.

We ran the same set of experiments with the defender using the DQN algorithm. The gradient of the {\tt BRS} function had to be slightly changed for the DQN algorithm. In general, we found the DQN results to be consistent with the PPO findings; DQN results are given in detail in the supplementary material. Overall, our experiments showed that adding reward shaping to the tag task (especially for Att-H) and to the boundary task (especially for Att-E) considerably improved the defender's rewards convergence time without degrading its peformance, indicating that our results support Hypothesis 1.

{\bf Hypothesis 2.} {\em When an agent has to learn multiple sub-tasks as part of its overall task, reward shaping for the more difficult tasks results in higher improvements in the agent's performance than reward shaping for the less difficult tasks.} A noteworthy observation in Figures~\ref{fig:ppo_training_score}(a)-(d) is that against Att-E, using {\tt BRS} improves the learning time over using {\tt TRS}, but against Att-H, this relationship between the learning times gets reversed. With {\tt BRS} the defender gets negative rewards for getting closer to the boundary, and consequently, learns to play in a more confined area. Against the fixed trajectory Att-E that moves close to the middle of the field, this culminates in more tagging opportunities, But against the more dexterous Att-H, playing in a smaller area away from boundaries leaves more unguarded space for the attacker to get through. Consequently, against Att-H, targeting the attacker more aggressively using {\tt TRS} yields faster learning times for the defender. To further validate this hypothesis we performed another experiment to make the tag task more difficult by changing the attacker from easy to hard while keeping the boundary learning task (size of the field) unchanged. Comparing Figures~\ref{fig:ppo_training_score} (a) and (b), we see that increasing the difficulty of the tagging sub-task Figure (b) results in agents trained using {\tt TRS} converge quicker and achieve higher scores compared to both {\tt SR} or {\tt BRS} in Figure (b). This shows that when the attacker goes from easy to hard (tag task gets harder), the training time with {\tt BRS} increases, while the training time using {\tt TRS} decreases as compared to {\tt SR}.

One anomaly we observe within the PPO algorithms is that while combining tag and boundary reward in the {\tt BTRS} case against Att-H, we are getting poorer scores than with {\tt TRS} or {\tt BRS} alone. A closer observation of the defender's moves showed that this was due to the tag reward interfering with the boundary reward and causing the defender to lose points by going out-of-bounds often. Additional experiments showed that for, increasing the boundary reward's weight to about $2$ times the tag reward could address this problem with {\tt BTRS}.

In summary, we see from these experiments that reward shaping for the harder boundary task gives more improvements than for the easier tag task, supporting Hypothesis 2 for PPO algorithms. Similar to the evaluation of Hypothesis 1 we verified these findings with DQN algorithms; the supplementary material has details of the DQN results. 

{\bf Hypothesis 3.} {\em Reward shaping can be used to encode additional objective functions like energy conservation without significantly diminishing the player's performance in the game.} A successful validation of Hypothesis 3 would show that when energy reward shaping is used, the agent chooses actions with lower energy requirements (e.g., stopping or maintaining speed and bearing between successive actions) more often than higher energy actions (e.g., changing speed or bearing). The energy reward shaping function, {\tt EFF},  has a limited effect against Att-E as stopping along the attacker's predefined path is already the optimal strategy that is converged on by all reward shaping functions expect {\tt 2BTRS}, as seen in Figure~\ref{fig:ppo_training_score} (e). The significance of {\tt EFF} is more evident against Att-H in Figure~\ref{fig:ppo_training_score} (f) where for most of the reward shaping functions the defender rarely uses the low energy stop or constant speed actions (white square at Speed: $0.0$ and Heading: $0.0$, light squares at most other headings); {\tt BRS} and {\tt 2BRS} prefer to stop more frequently, but they perform poorly against Att-H as shown in Figures~\ref{fig:ppo_training_score}(b) and (d). Only when the defender uses energy reward shaping {\tt EFF}, the stop action is chosen very often (dark square at EFF-Speed: $0.0$, light squares at all other headings). Sacrificing mobility for energy efficiency evidently reduces the defender's score by around $50-60\%$ (graph in supplementary material) with {\tt EFF} as compared to other reward shaping functions with full mobility. But analysis of the attacker and defender trajectories in the game show that by using {\tt EFF} reward shaping versus an Att-H attacker, the defender can learn to strategically position itself very near to its flag - the attacker avoids getting tagged on its approach trajectory to grab the flag, but the defender is able to tag the attacker on its exit trajectory and retrieve the grabbed flag while barely moving from its learned stationary position. {\footnote{We found that adjusting the {\tt EFF} reward function hyper-parameters in Equation~\ref{eqn:reward_shaping_enrg} can reduce the defender's propensity to remain stationary, and, consequently improve its scores, but at the expense of spending more energy.}}. Overall, the defender's behavior with {\tt EFF} reward shaping support our claim in Hypothesis 3 that additional objectives of a player like reducing energy consumption during playing can be controlled via a reward-shaping function without significantly degrading the player's performance. Reward shaping with {\tt EFF} within the DQN algorithm also supports these claims; details are in the supplementary material.

\begin{figure}
    \centering
   
    \includegraphics[width = 3.5in]{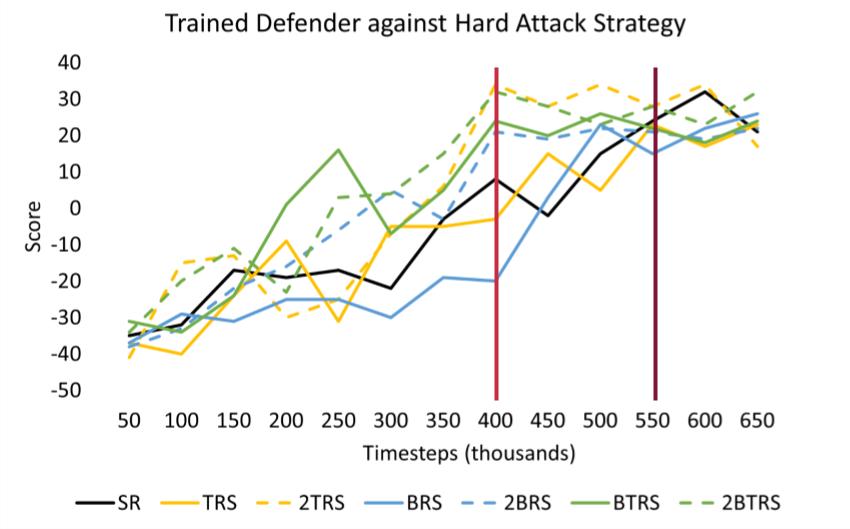} 
    
    \caption{Training times for a defender policy playing against Att-E using $1X$ (solid lines) and $2X$ (dashed lines) gradients with {\tt TRS}, {\tt BRS} and {\tt BTRS} reward shaping functions. The red and magenta vertical lines show the number of time-steps at which the defender scores converge with $2X$ and $1X$ gradients respectively.}
    \label{fig:medresults}
\end{figure}

{\bf Hypothesis 4.} {\em Hyper-parameter tuning of the reward function can guarantee faster convergence to higher rewards depending on the opponent's behavior.} To validate Hypothesis 4, we changed the gradients of {\tt TRS} and {\tt BRS} functions ($\kappa_2, \kappa_4$ in Equation~\ref{eqn:reward_shaping_bound} and $\omega_2, \omega_4$ in Equation~\ref{eqn:reward_shaping_tag}) to twice their original gradients. Figure~\ref{fig:medresults} shows that defender scores against Att-H (dashed lines) for all the reward shaping functions with $2X$ gradient converge to a score between $20-30$ within $400$K training time-steps (marked with a red vertical line). In contrast, while using $1X$ gradient for reward shaping functions, defender scores against Att-H (solid lines) take $550$K training time-steps (solid magenta vertical line) to converge to comparable scores. This corresponds to a significant improvement of $150$K training time-steps for $2X$ gradient. Similar experiments of changing the reward shaping function gradients from $1X$ to $2X$ against Att-E yielded positive, but smaller improvements of $5$K-$10$K fewer training time-steps with $2X$ versus $1X$, for all the reward shaping functions. Finally, changing the gradient from $1X$ to $0.5X$ resulted it poorer performance against both Att-E and Att-H; it increased the number of training time-steps required for scores to converge as well as degraded the final converged scores. Overall, from these experiments, we see that a player's performance in terms of score and learning times can be significantly affected by hyper-parameter tuning of the gradient of the reward shaping function. Also, the optimal hyper-parameters are a function of the opponent strategy and they have to be re-tuned for different opponent strategies. Overall, these findings show that our Hypothesis 4 is supported. 

\begin{figure}
    \centering
    \begin{tabular}{c}
    \includegraphics[width=3.5in]{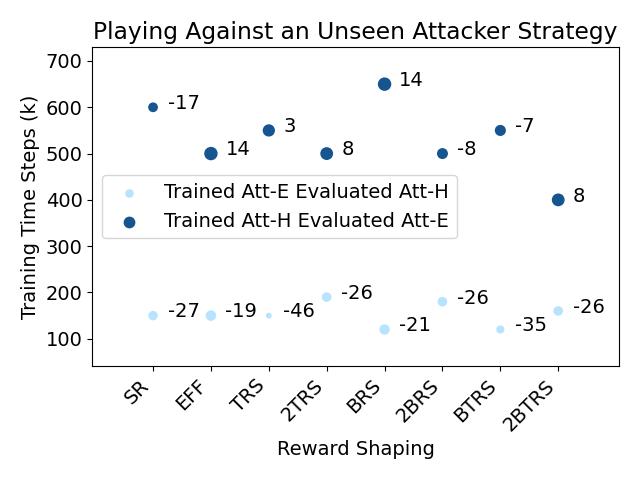} \\
    \includegraphics[width=3.5in]{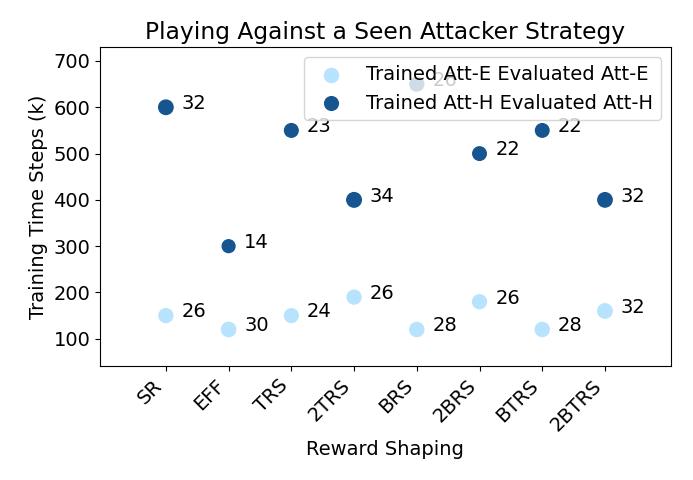} \\
    \end{tabular}
    \caption{Defender training times (y-axis) and scores (shown next to circle) using different reward shaping functions while playing against an unseen attacker strategy (top), and same attacker strategy as the one trained against (bottom).}
    \label{fig:ppo_scatterplot}
\end{figure}

\begin{figure*}[t]
    \centering
    \begin{tabular}{ccc}
        \hspace*{-0.2in}        
        \includegraphics[width=2.5in]{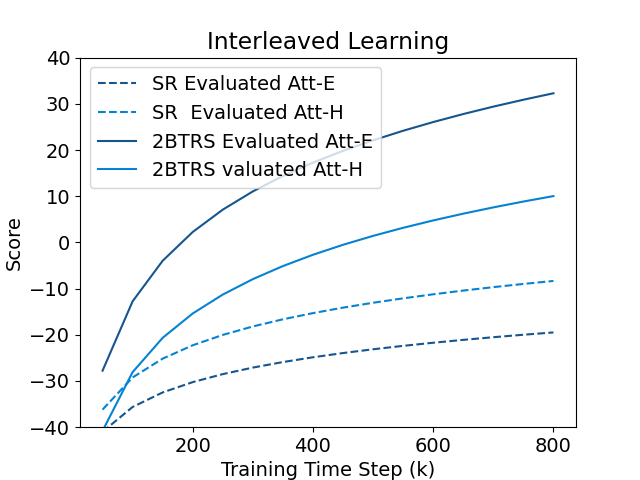}& 
        \hspace{-0.3in}
        \includegraphics[width=2.5in]{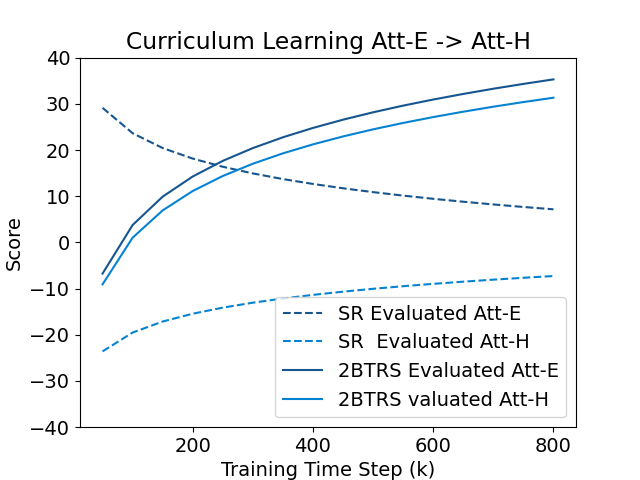} &
        \hspace{-0.3in}
        \includegraphics[width=2.5in]{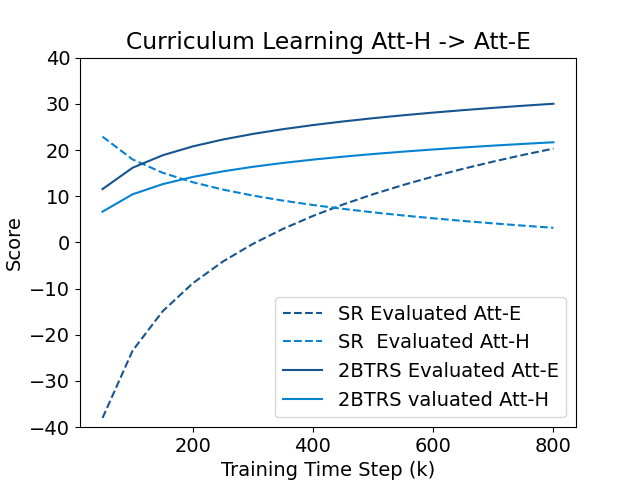} \\
        (a) & (b) & (c)
    \end{tabular}
    \caption{Training curves for the defender with {\tt SR} and {\tt 2BTRS)} with (a) interleaved training: the attacker (Att-E or Att-H) trained against is picked randomly at the start of every training episode, (b) curriculum learning (E $\rightarrow H)$: training first against Att-E up to convergence followed by training against Att-H, (c) training first against Att-H up to convergence and then against Att-E.}
    \label{fig:gen_policy_curriculum_learning}
\end{figure*}

{\bf Hypothesis 5.} {\em Reward shaping improves learning generalizable policies.} The main objective of Hypothesis 5 is to evaluate if reward shaping can improve learning policies that can perform well against different skill levels of attackers. We performed two sets of experiments to evaluate this hypothesis. In the first experiment, we evaluated policy transfer, that is, if a defender policy trained against one attacker type could continue to play effectively if it encountered the other attacker type during evaluation. The results from the experiment are given in Figure~\ref{fig:ppo_scatterplot}. We see in Figure~\ref{fig:ppo_scatterplot} (top) that when a defender trained against one attacker strategy (Att-E or Att-H) but played against a different attacker strategy than the one trained on, it scores diminished significantly. For comparison, the defender scores while playing against the same attacker strategy it had been trained with are given in Figure~\ref{fig:ppo_scatterplot}(bottom). The results also show that reward shaping can provide minimal improvement on these transfer learning tasks. 

In order to create more general policies, we looked at two methods, interleaved and curriculum learning, as discussed in Section~\ref{sec:reward_shaping}. For training in interleaved learning, at each training episode we randomly picked the attacker type, Att-E or Att-H, the defender would train against. The results can be seen in Figure~\ref{fig:gen_policy_curriculum_learning}(a). Without reward shaping the generalized learning tasks don't converge within $800$K steps (dashed lines), which is likely attributed to learning game features other than tagging range or game boundaries within the game context. On the other hand, generalized learning using {\tt 2BTRS} reward shaping results in the convergence of a policy against both Att-E and Att-H (solid lines), comparable to just training a policy against one of these strategies. This more efficient training could be attributed to reward shaping, which allows the agent to learn the boundary game context when playing against the Att-E strategy and then gain more context on the tagging range when playing against the Att-H strategy.  

For training with curriculum learning, a defender was initially trained against Att-E for $150$K time-steps and then the Att-E trained policy was further trained against Att-H up to $800$K time-steps. The training curve for training the Att-E trained policy against Att-H is shown in Figure~\ref{fig:gen_policy_curriculum_learning}(b). We see that using sparse rewards results in learning loss (dashed lines) - the defender' score against Att-E decreases as it trains more against Att-H and the policy does not generalize to play against both Att-E and Att-H. In contrast, with reward shaping with {\tt 2BTRS} (solid lines), we see that the defender's score improves against Att-H without learning loss against Att-E. We repeated the curriculum learning experiment while first training the defender against Att-H for $550$K time-steps followed by training against Att-E up to $800$K time-steps. The results  in Figure~\ref{fig:gen_policy_curriculum_learning}(c) show that {\tt SR} faces learning loss as its score against Att-H decreases (dashed lines), but with {\tt 2BTRS} reward shaping the defender can generalize to play well against both strategies. These results show that reward shaping within curriculum learning leads to a defender policy that can learn to play against a new strategy without forgetting to play against the initial strategy it had learned to play against, than with only sparse rewards. In summary, we show that reward shaping improves learning a generalizable policy that can continue to perform well if the opponent dynamically changes its strategy during game-play, supporting Hypothesis 5.

\section{Conclusions and Future Directions}
Our main contribution in this paper showed that reward shaping can be used as an effective yet computationally frugal means to improve a player's performance and training times in a CTF RTS game. Additionally, reward shaping could identify which sub-tasks have a bigger impact on a player's performance and wins in a game, and could also be used to encode more important, secondary performance objectives such as reduced energy expenditure into a player's behavior. In this work, we have hand-crafted different reward-shaping functions based on common sense and human intuition. An interesting future problem would be to use a learning algorithm to determine and fine-tune suitable reward shaping functions along the directions in~\cite{mguni2021learning}. Another related problem is to automatically determine where or for which sub-tasks in a game do the reward functions have the most impact on a player's performance. Since the reward shaping function tuning is a non-trivial task and has to be done in parallel to the main task of learning to play the game, this would provide us a recipe for suitably allocating computing resources for reward shaping tasks.
Future work exploring these concepts would be to design a general policy that leverages a learned hierarchical policy and then compare the compute time correlates to a pure general learning approach. 
\\

{\bf Credit Authorship Statement.} Dasgupta was responsible for conceptualizing and supervising the research and writing Sections I-IV of the paper. Kliem was responsible for developing the software code, running the experiments and presenting the research results via graphs and other visualizations. Sections V-VI and the supplementary material in the paper were written jointly by both authors.

{\bf Acknowledgements.} This research was supported by the Office of Naval Research through an Naval Research Laboratory $6.1$ Base Funding project grant titled 'Game Theoretic Machine Learning with Defense Applications'.

\bibliographystyle{IEEEtran}
\bibliography{refs}

\clearpage
\section*{Appendix}

\begin{figure*}[t]
    \centering
    \begin{tabular}{ccc}
        \hspace*{-0.2in}        
        \includegraphics[width=2.5in]{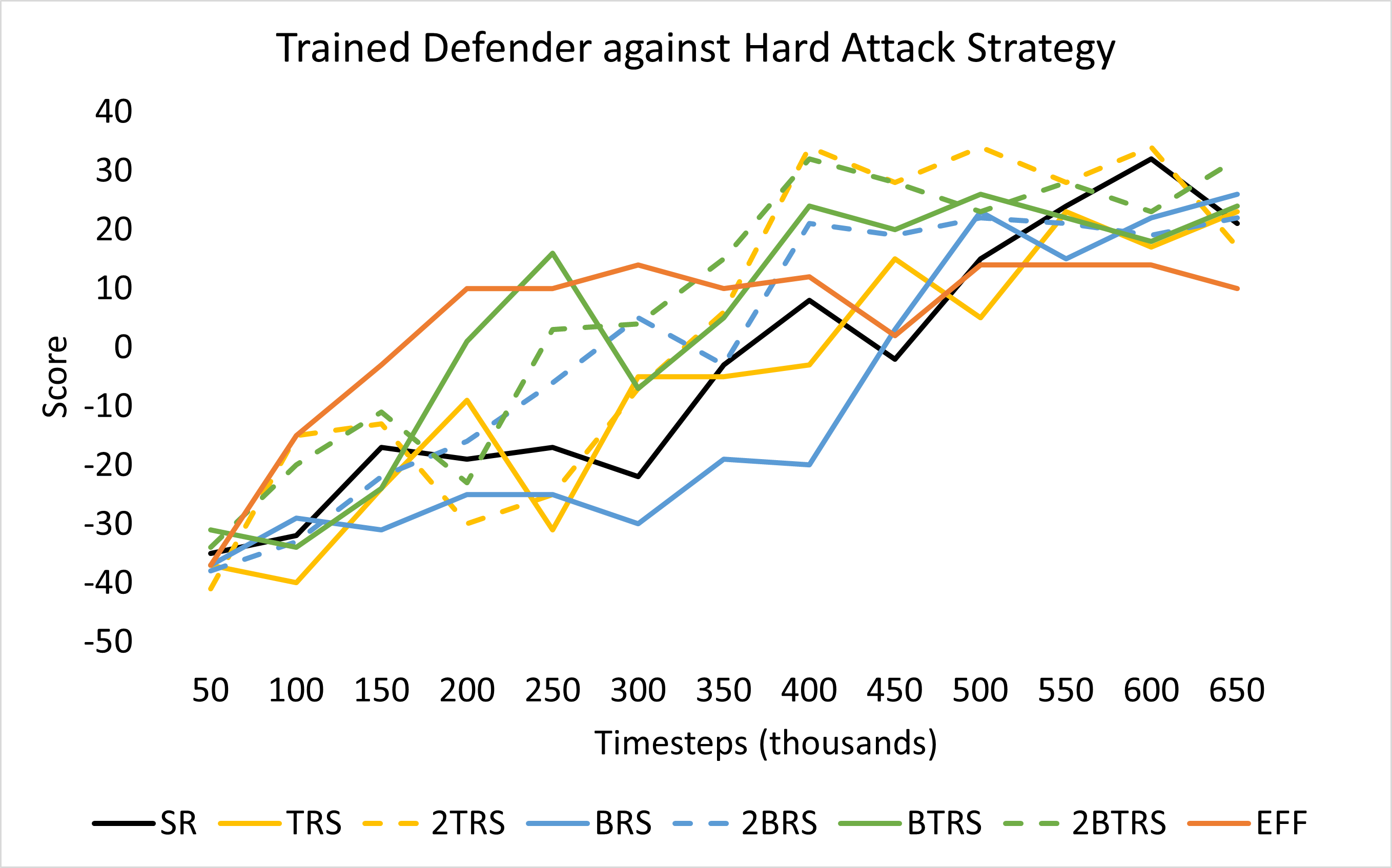}& 
        \hspace{-0.3in}
        \includegraphics[width=2.5in]{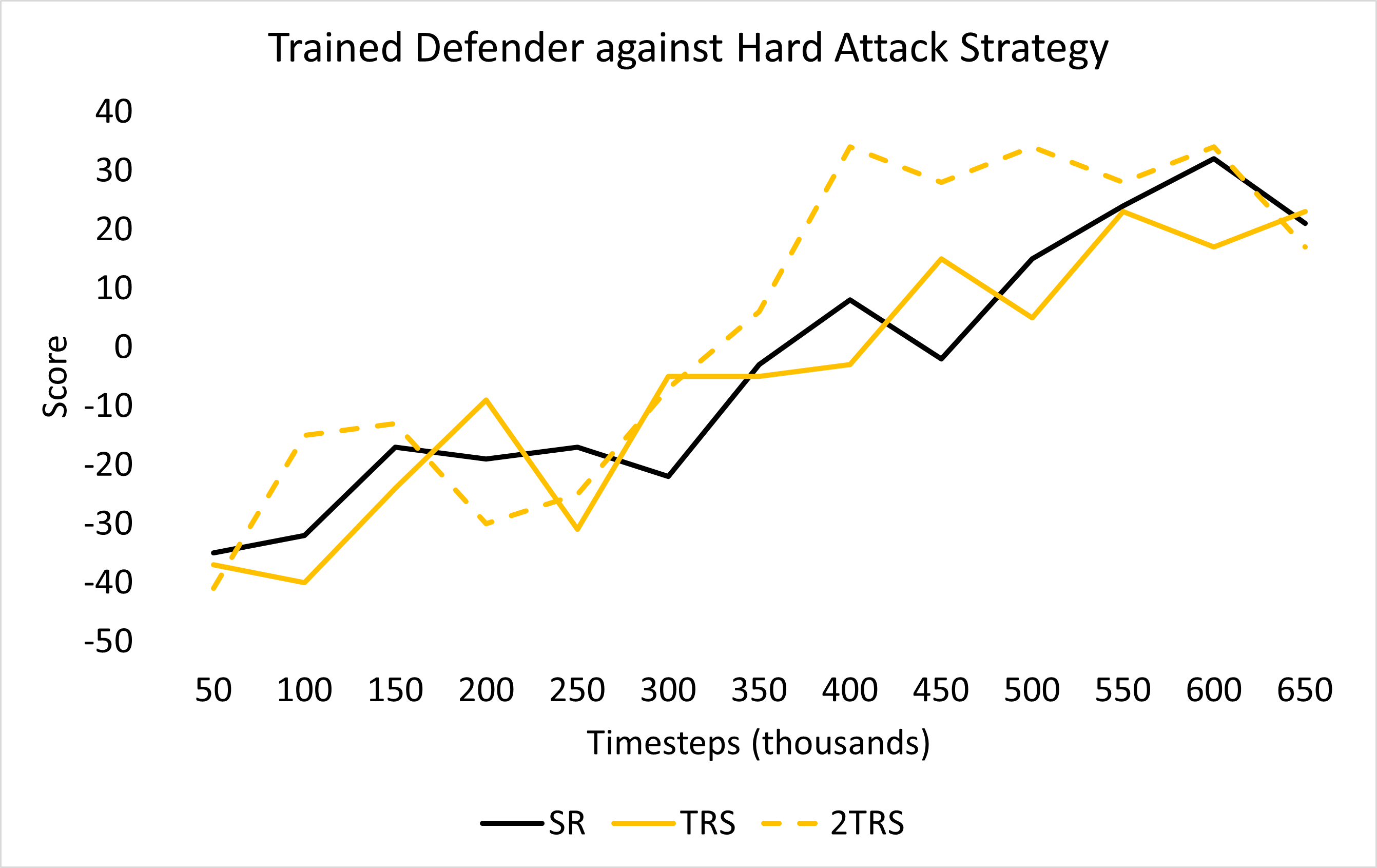} &
        \hspace{-0.3in}
        \includegraphics[width=2.5in]{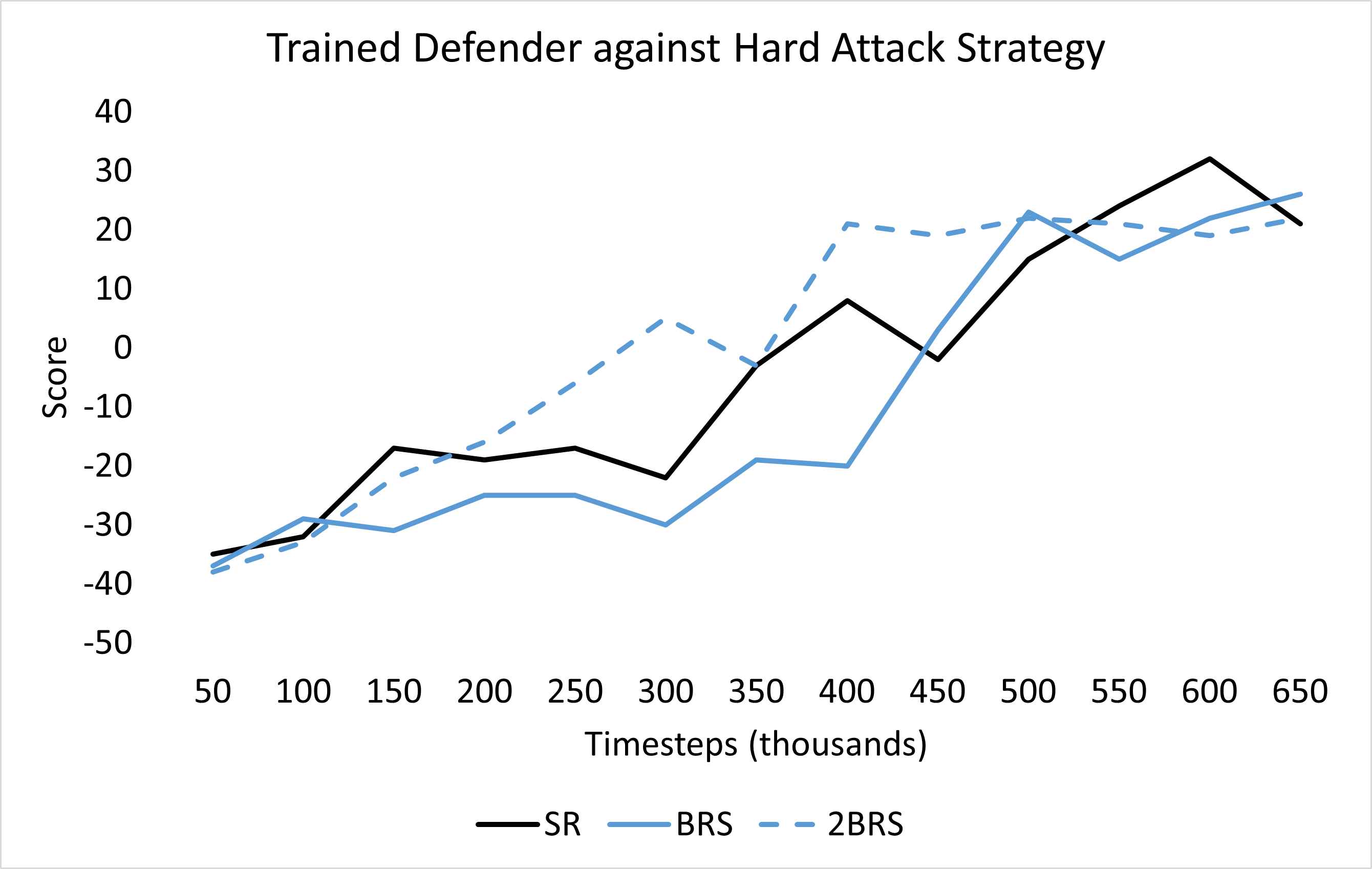} \\
        (a) & (b) & (c) \\
        \includegraphics[width=2.5in]{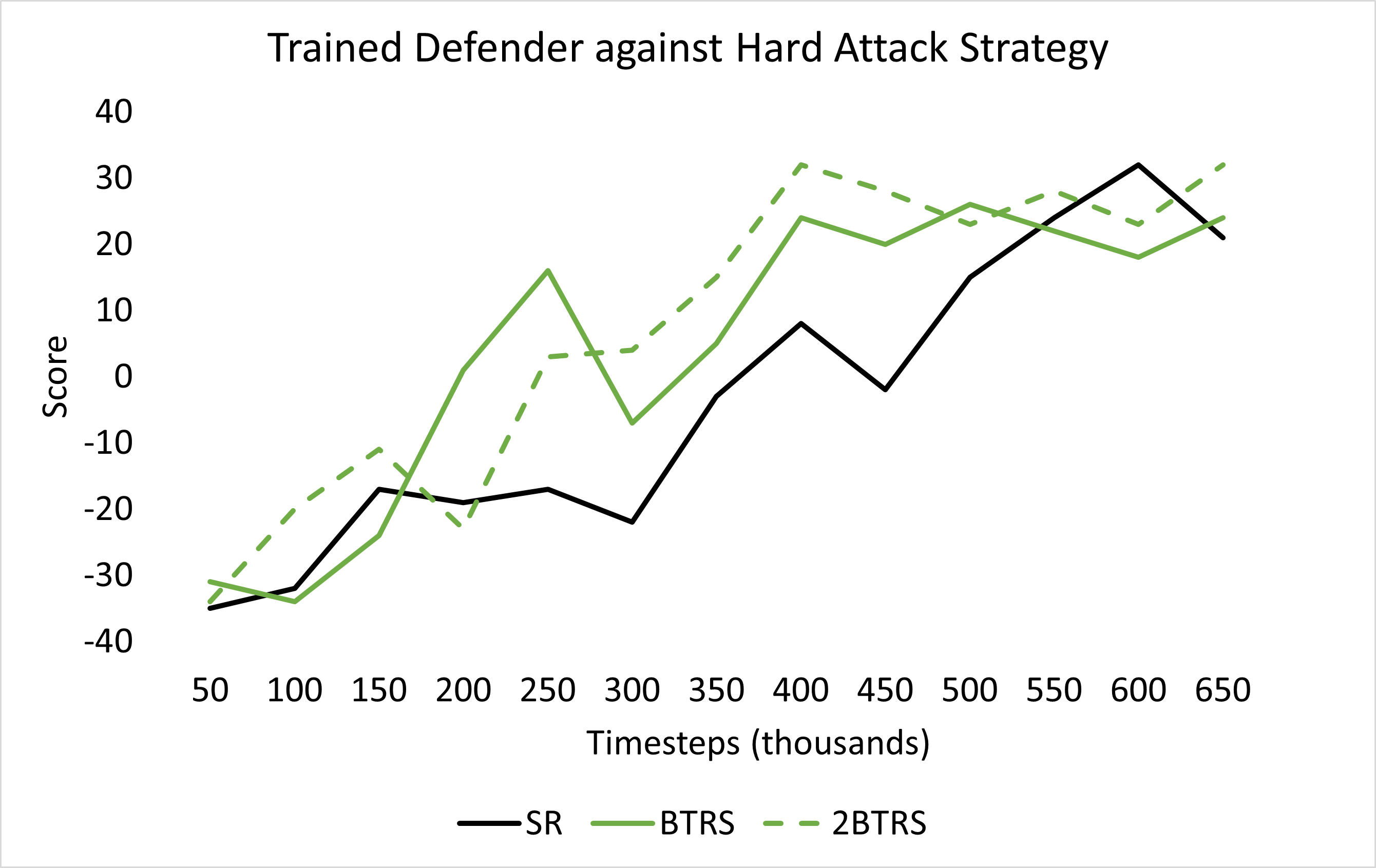}& 
        \hspace{-0.3in}
        \includegraphics[width=2.5in]{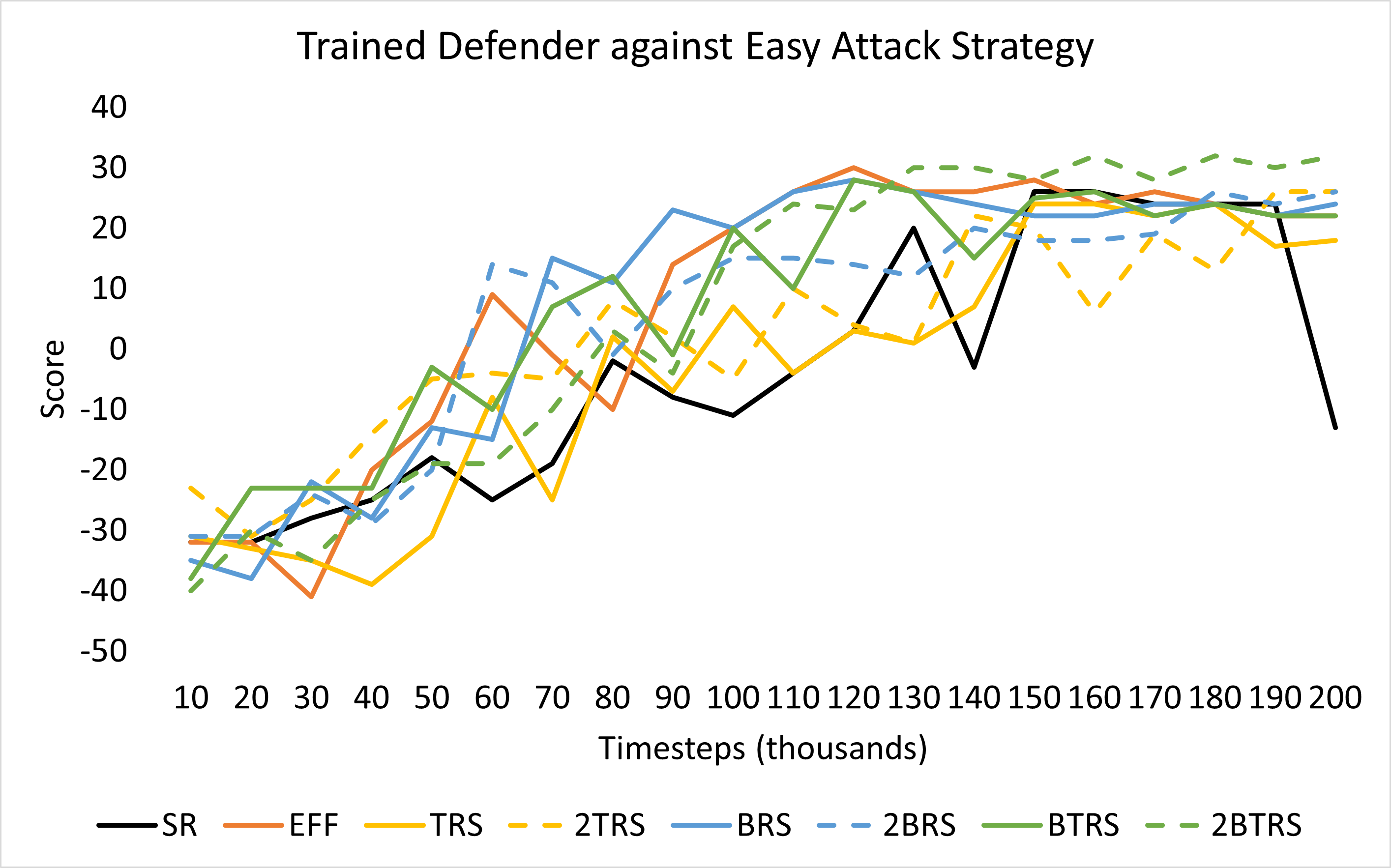} &
        \hspace{-0.3in}
        \includegraphics[width=2.5in]{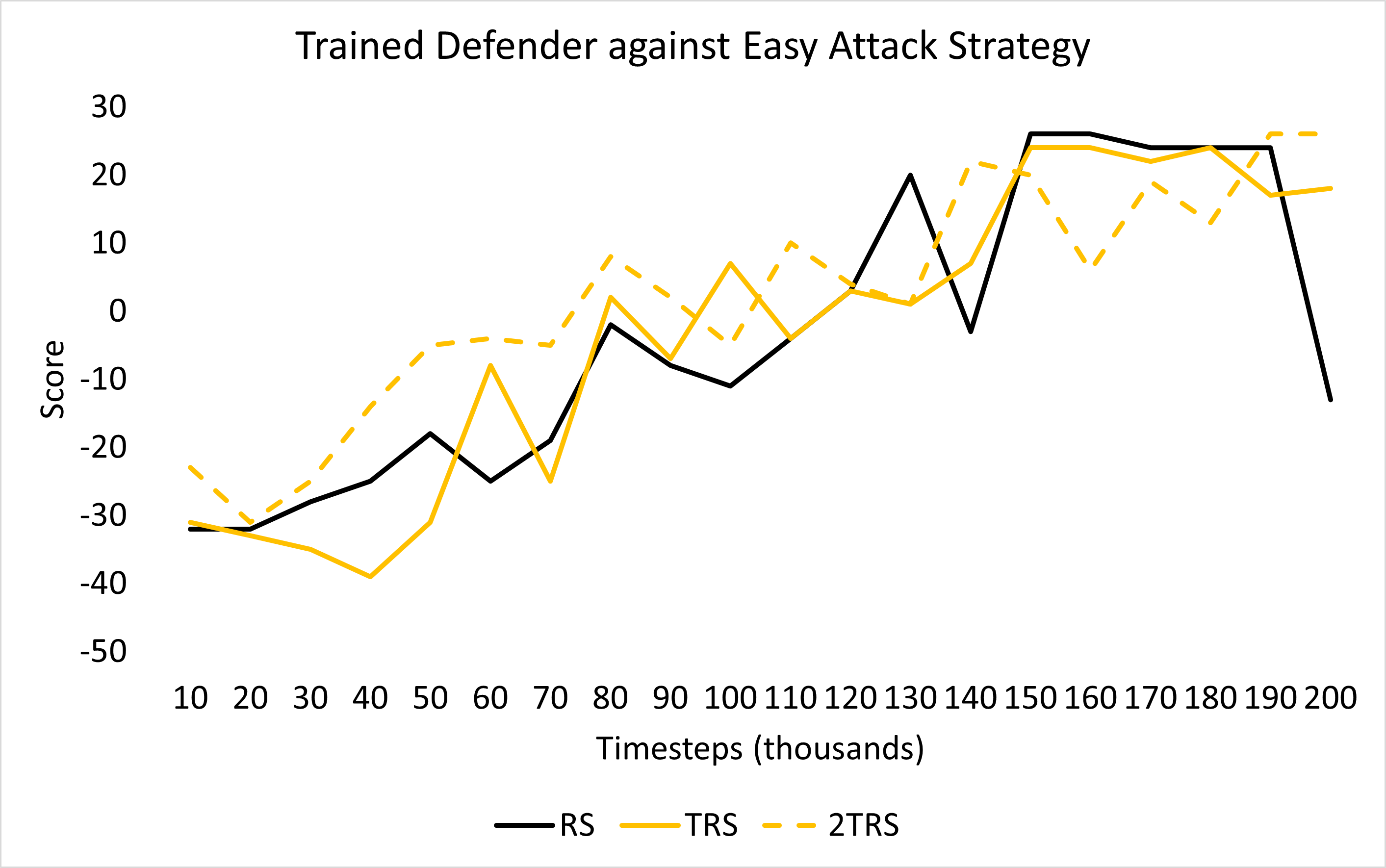}
        \\
        (d) & (e) & (f) \\
        \includegraphics[width=2.5in]{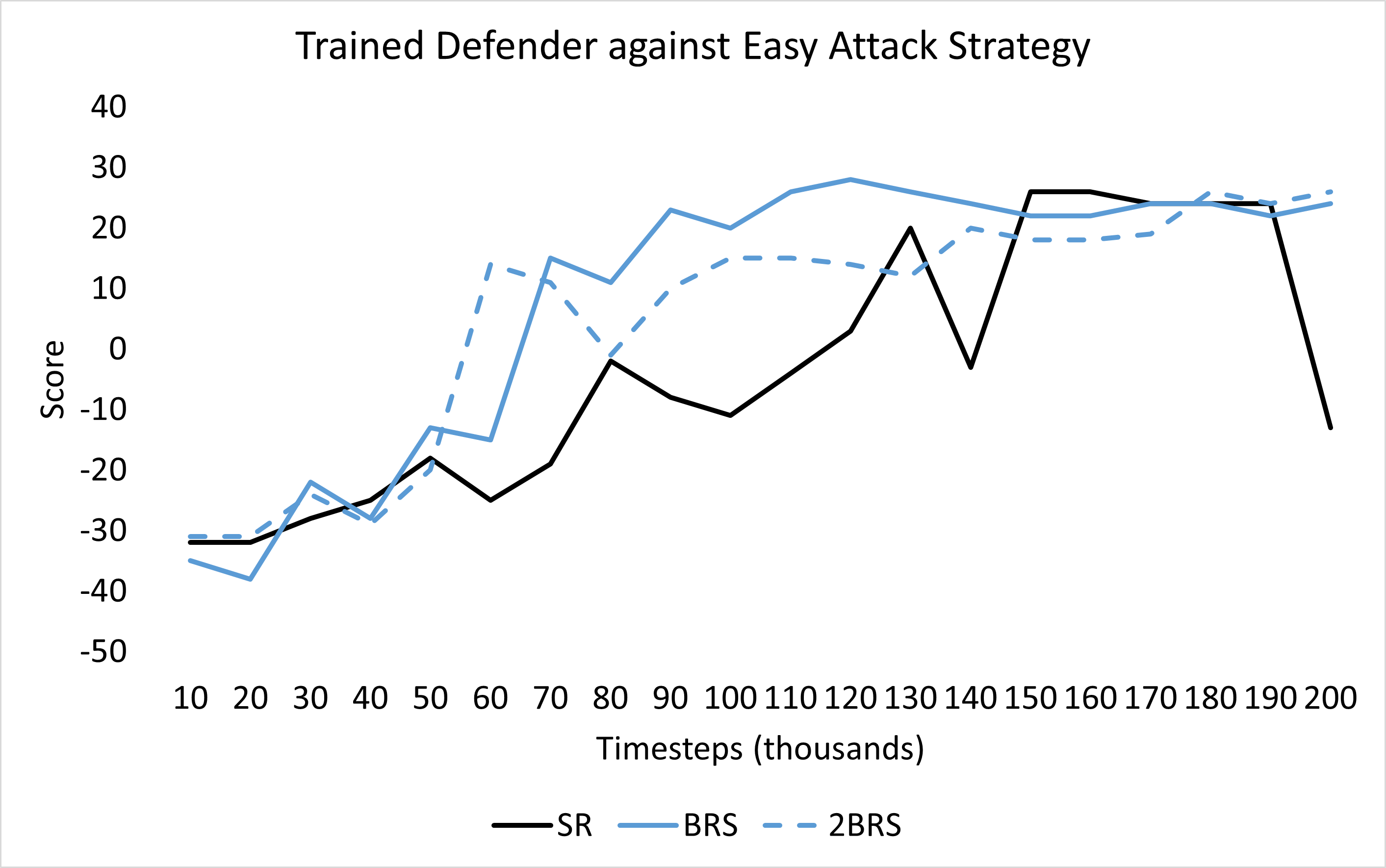}& 
        \hspace{-0.3in}
        \includegraphics[width=2.5in]{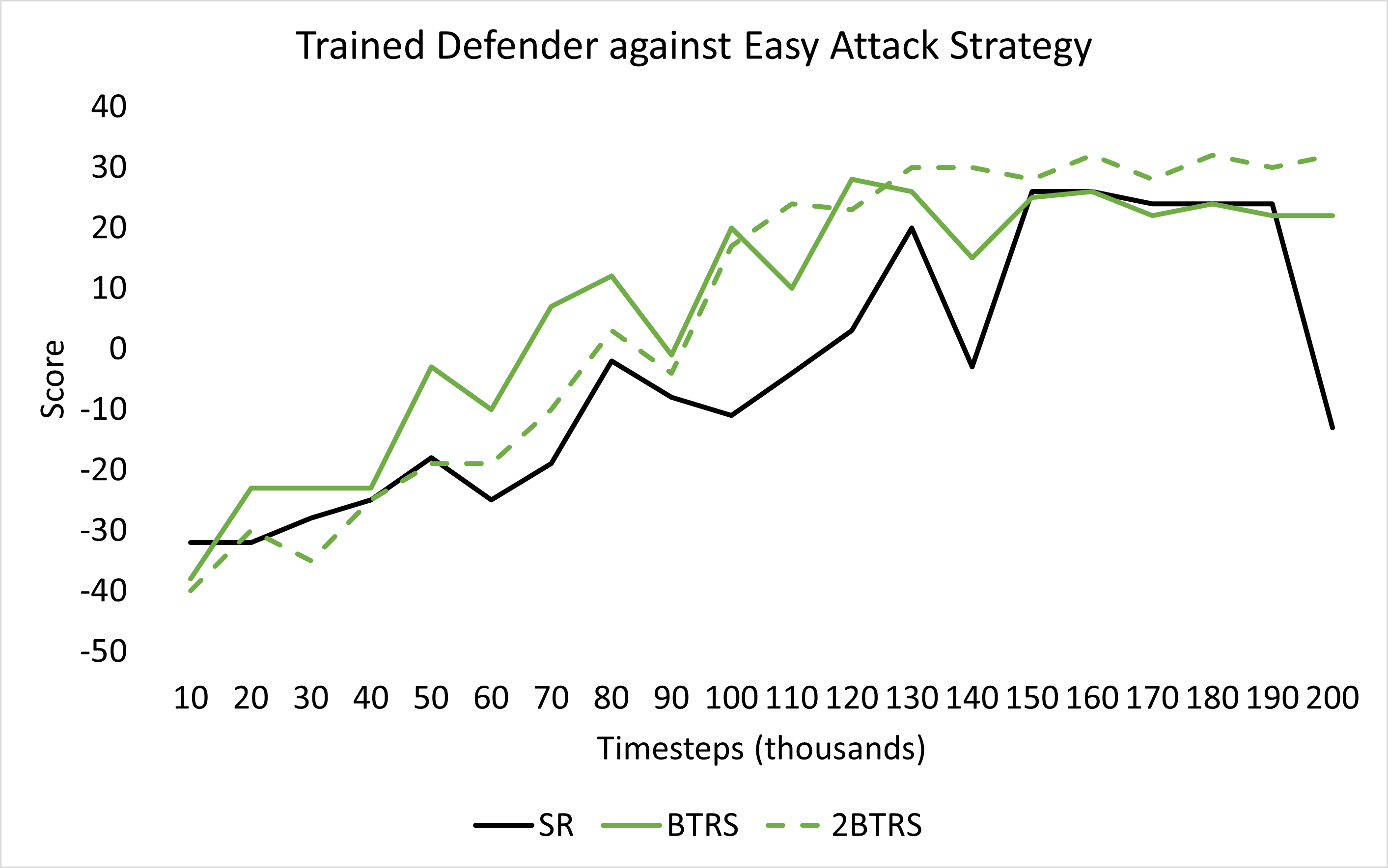} &
        \hspace{-0.3in}
        \includegraphics[width=2.5in]{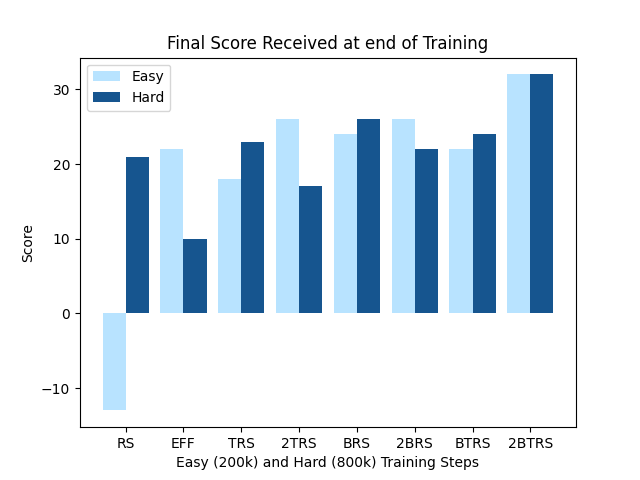} \\
        (g) & (h) & (i)
    \end{tabular}
    \caption{Training times for different gradients of the TRS, BRS and BTRS reward shaping functions (a-d) Att-H and (e-h) Att-E. Figure (i) bar graph representing the final scores achieved at the end of training 200k timesteps against Att-E and 800k timesteps against Att-H.}
    \label{fig:gen_policy_curriculum_learning}
\end{figure*}

\begin{table}
\centering
\begin{tabular}{|c|l|}
\hline
Notation & Description\\
\hline
\hline
$\theta_p$ & Player's heading\\
\hline
$\{(d_{o_j}, \hat{\theta}_{o_j}, \theta_{o_j})\}$ & Distance, angle and heading to opponent $j$\\
\hline
$\{(d_{p_j}, \hat{\theta}_{p_j}, \theta_{p_j})\}$ & Distance, angle and heading to teammate $j$\\
\hline
($d_{flag,o}, \hat{\theta}_{flag, o})$& Distance, angle to opponent's flag\\
\hline
($d_{flag,p}, \hat{\theta}_{flag, p})$& Distance and angle to player's (self) flag\\
\hline
$\mathbf{D}_{bounds} = \{d_{upper}, d_{lower},$ & Distances to upper, lower, left and right \\
$d_{left}, d_{right}\}$ & (lines) boundaries of playing field \\
\hline
\end{tabular}
\caption{Feature space derived from observations used for training the RL algorithm of a player in the Aquaticus CTF Game.}
\label{table:feature_space}
\end{table}

\subsection*{Event State Sets}
We define the following state sets corresponding to game events where players score or lose points:
\begin{itemize}
    \item {\em Tag set} is the set of states at which players can tag each other, before an attacker has grabbed the defender's flag. Formally, it is defined as $S^{tag} = \{s: s = (s^{def}, s^{att}) \in S \wedge |\mathbf{x}^{def} - \mathbf{x}^{att}| \leq D^{tag}\}$ (before attacker grabs flag). A defender tag happens if both players are in the defender's zone in a tag state, otherwise it is an attacker tag.
    \item {\em Retrieval Tag Set} is similar to a tag set, except it happens only after the attacker has grabbed the defender's flag. Formally, it is defined as $S^{ret} = \{s: s = (s^{def}, s^{att}) \in S \wedge |\mathbf{x}^{def} - \mathbf{x}^{att}| \leq D^{tag}\}$ (after attacker grabs flag). The same conditions for defender and attacker tag as in the tag set apply to retrieval tag states as well.
    \item {\em Grab Set} is the set of states at which an attacker grabs the defender's flag from the defender's base. Formally, $S^{grb} = \{s: s = (s^{def}, s^{att}) \in S \wedge |\mathbf{x}^{att} - \mathbf{x}^{dflg}| \leq D^{grab}\}$.
    \item {\em Capture Set} is the set of states at which an attacker captures the defender's flag by bringing it back to its own base. Formally, $S^{grb} = \{s: s = (s^{def}, s^{att}) \in S \wedge |\mathbf{x}^{att} - \mathbf{x}^{abase}| \leq D^{cap}\}$.
    \item {\em Out-of-Bounds (OOB) Set} is the set of states at which a player is outside the playing field. Formally, the defender OOB set is defined as $S^{oob, def} = \{s: s = (s^{def}, s^{att}) \in S \wedge \mathbf{x}^{def} \notin \mathbf{X}^{field}\}$. The attacker OOB set $S^{oob, att}$ is defined commensurately.
\end{itemize}
The game event set $S^{EV} = \{S^{tag}, S^{ret}, S^{grb}, S^{cap}, S^{oob}\}$ gives a collection of the above state sets.

\subsection{Attacker and Defender Sparse Reward Functions}
The attacker and defender sparse reward functions specify the external reward component of their respective reward functions given in Equation~\ref{eqn:reward_def_ext_int}. They are obtained by multiplying the points obtained at game event states $S^{EV}$ by a constant scaling factor $C_{ext}$. The different attacker and defender sparse reward functions for $C_{ext} = 50$ (used in our experiments) are given below:
\[
    R^{tag, def}= 
\begin{cases}
	100, & \text{if } {|\mathbf{x}^{att}- \mathbf{x}^{def}|\leq D^{tag}} \text{ and} \\
        & \mathbf{x}^{att}, \mathbf{x}^{def} \in Z^{def}, \quad \text{(before grab)}\\
    0, & \text{otherwise}
\end{cases}
\]

\[
    R^{ret, def}= 
\begin{cases}
	50, & \text{if } {|\mathbf{x}^{att}- \mathbf{x}^{def}|\leq D^{tag}}\text{ and} \\
        & \mathbf{x}^{att}, \mathbf{x}^{def} \in Z^{def}, \quad \text{(after grab)}\\
    0, & \text{otherwise}
\end{cases}
\]

\[
    R^{grab, att}= 
\begin{cases}
    50,& \text{if } {|\mathbf{x}^{att} - \mathbf{x}^{dflg}| \leq D^{grab}}\\
	0, & \text{otherwise}
\end{cases}
\]

\[
    R^{cap, att}= 
\begin{cases}
    100,& \text{if } {|\mathbf{x}^{att} - \mathbf{x}^{abase}| \leq D^{cap}},\quad \text{(after grab)}\\
        & \text{ where } \mathbf{x}^{abase} \in \mathbf{X}^{abase}\\
	0, & \text{otherwise}
\end{cases}
\]

\[
    R^{oob, def} = 
\begin{cases}
    -100,& \text{if  player out-of-bounds}\\    
     0, & \text{otherwise}
\end{cases}
\]

\omitit{The total external reward to the defender is given by:
\begin{eqnarray}
R^{def}_{ext} & = & R^{tag, def} + R^{ret, def} - R^{grab, att} - R^{cap, att} \nonumber \\
 & & + R^{oob, def}
\label{eqn:sparse_reward_def}
\end{eqnarray}
}

\subsection{Deep Q-Network and Related Experiments}

\begin{figure}
    \centering
    \begin{tabular}{cc}
    \hspace{-0.1in}
    \includegraphics[width=1.7in]{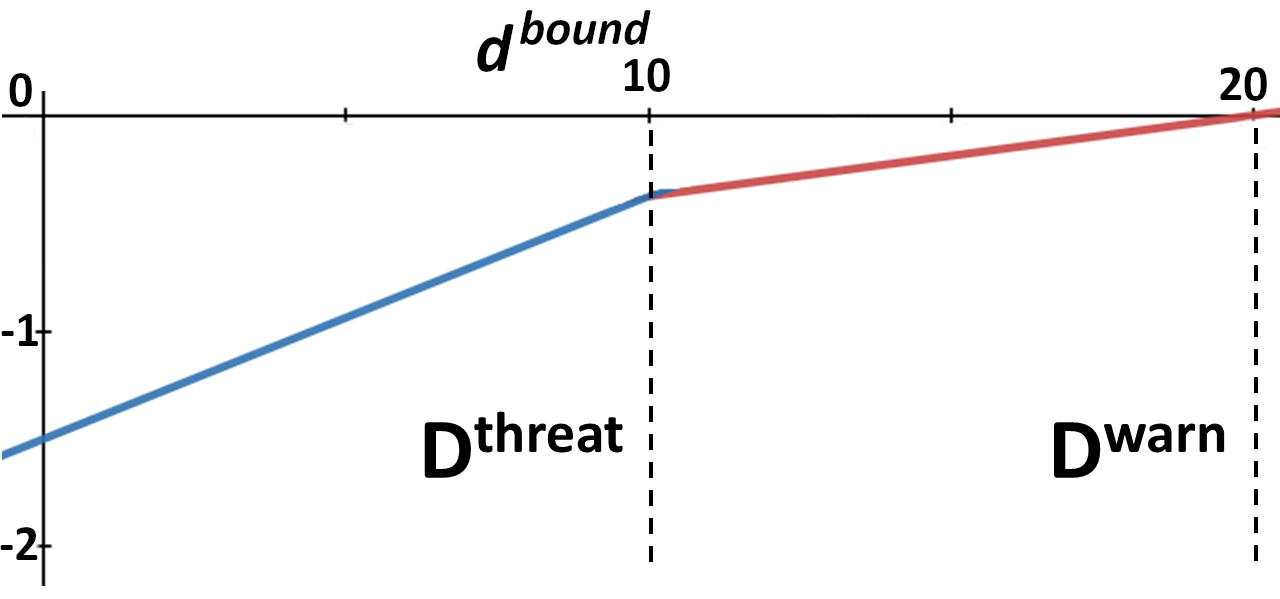} &
    \hspace{-0.15in}
    \includegraphics[width=1.7in]{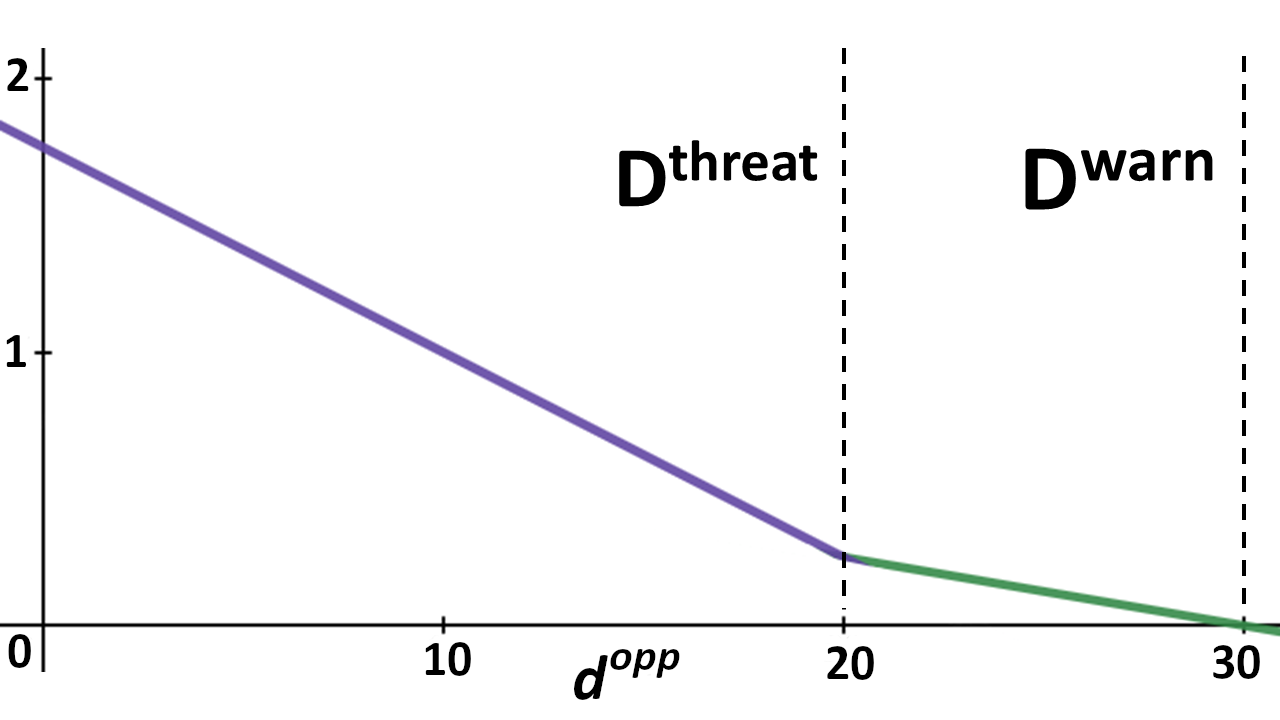} \\
    {\small (a)} & {\small (b)}\\
    \end{tabular}
    \caption{Example graphs for (a) boundary and (b) tag reward shaping functions.}
    \label{fig:rewardshapingdqnfig}
\end{figure}

{\bf DQN Experimental Settings.} The same game settings used for the DQN results are the same as the PPO results except where explicitly stated. The parameters used in the reward shaping equations, Equations~\ref{eqn:reward_shaping_bound}-~\ref{eqn:reward_shaping_enrg} are $\kappa_1=-0.75$, $\kappa_2=0.0375$, $\kappa_3=-1.5$, $\kappa_4=0.1125$, $\omega_1=0.75$, $\omega_2=0.025$, $\omega_3=1.75$, $\omega_4=0.075$, $\mu_1=0.5$, $\mu_2=0.4$, and, $\mu_3=-0.5$.

We have used Stable-Baselines 2 library~\cite{stable-baselines} for the DQN algorithm implementation. The policy network of the DQN has a multi-layer perceptron with $2$ layers of $64$ perceptrons each. DQN training hyper-parameters are: discount factor, $\gamma = 0.99$, learning rate, $\alpha = 0.0005$, and buffer size$=5000$. Each DQN training was done for a minimum of $600$K time-steps, corresponding to roughly $16$ hours of clock time for each training.

{\bf DQN Hypothesis 1.} To validate Hypothesis 1, we trained a defender against an easy attacker while using sparse rewards only ({\tt SR}), and reward shaping functions for tag ({\tt TRS}), boundary ({\tt BRS}), and a combination of boundary and tag ({\tt BTRS}). Our results are summarized in Table~\ref{table:trained_e_vs_e}. Each sub-table gives the attacker's performance for one of these reward function settings. The performance of the attacker in terms of its main events, grab and capture, and of the defender, in terms of its main events, tagging the attacker, retrieving the flag from the attacker and getting tagged are recorded at intervals of $50,000$ training steps. When the defender plays optimally, the attacker should get $0$ grabs and captures, while the defender should get most possible tags, $0$ retrieval tags (as it tags the attacker before attacker grabs flag), and $0$ defender tags (never get tagged by attacker or leave the playing field). Highlighted rows in Table~\ref{table:trained_e_vs_e} show the best performance for each of our reward function settings. We see that with sparse rewards (leftmost sub-table) the defender gets its best result at $400$K training steps when it gets $10$ tags ($6$ before attacker's flag grab, $4$ before attacker's flag capture) and never gets tagged itself. When we augment sparse rewards with tag reward shaping (second sub-table), the defender's rewards convergence time reduces by $50\%$ from $400$K to $200$K training steps, although its performance in terms of tags remains comparable to sparse rewards, albeit a bit lower in terms of defender getting tagged $4$ times due to leaving the playing field. A higher improvement over sparse reward comes with adding boundary reward shaping (third sub-table). It improves the reward convergence time by $75\%$ over sparse reward while increasing the number of before-grab tags to $9$. Finally, the best result comes with combining boundary and tag rewards - the reward convergence time improves by $87.5\%$ and performance improves to $11$ tags, all before the attacker is able to grab the flag even once, without the defender itself ever getting tagged. Overall, we see that adding reward shaping to the tag task and, more importantly, to the boundary task considerably improves the rewards convergence time and game-play quality of the defender. These results indicate that Hypothesis 1 is supported.

{\bf DQN Hypothesis 2.} For validating Hypothesis 2, note that in Table~\ref{table:trained_e_vs_e} learning the boundary task is more difficult and more important for the defender. Consequently, shaping the boundary reward gives a bigger improvement in performance than shaping tag reward. For further validation of Hypothesis 2 we did another experiment to make the tag task more difficult by changing the attacker from easy to hard while keeping the boundary learning task (size of the field) unchanged. In Table~\ref{table:trained_e_med_vs_e_med_score}, we report the defender's scores calculated using Equation~\ref{eqn:score_def} for this experiment. The rows of the sub-tables correspond to defenders trained against an easy attacker (Def-E) or hard attacker (Def-H), while the columns correspond to easy and hard attackers (Att-E and Att-H) that the defenders in the rows are evaluated to play against. The scores in Table~\ref{table:trained_e_med_vs_e_med_score} show that when the attacker goes from easy to hard (tag task gets harder), the performance of defender Def-E diminishes significantly. Also, as shown in the bottom sub-table (Def-H vs. Att-H), the defender is able to recuperate its performance to a score of $20$ at $100$K training steps indicating that the boundary task is a more difficult learning task even with the hard attacker and shaping its reward gives the most performance improvement. In summary, we see from these experiments that reward shaping for the harder boundary task gives more improvements than for the easier tag task, supporting Hypothesis 2. One anomaly we observe is that while combining tag and boundary reward in the BTRS case against Att-H, we are getting poorer scores than with reward shaping for tag (TRS) or boundary (BRS) alone. A closer observation of the defender's moves showed that this was due to the tag reward interfering with the boundary reward and causing the defender to lose points by going out-of-bounds often. Additional experiments showed that for the BTRS case, increasing the boundary reward's weight to about $8$ times the tag reward could address this issue, improving the Def-H defender's score to $8$ in $150$K training steps against Att-H.

\begin{figure}[t]
   \hspace*{-0.3in}
        \includegraphics[width=4.0in]{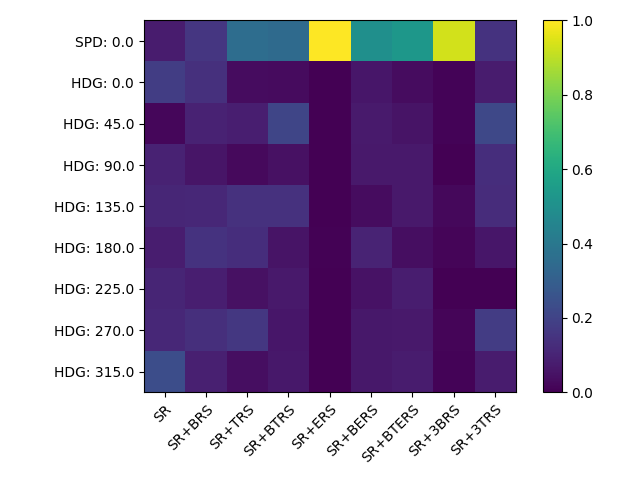}
    \caption{Action heat map showing actions taken by defender playing against Att-E with: (SR) sparse reward, (BTRS) bound and tag reward, (3BRS) boundary reward with $3 \times$gradient, (3TRS) tag reward shaping with $3 \times$gradient, (ERS) efficiency reward, (BERS) boundary with efficiency reward, (BTERS) boundary, tag with efficiency reward.}
    \label{fig:action_heat_maps_dqn}
\end{figure}

{\bf DQN Hypothesis 3.} Our Hypothesis 3 relates to evaluating if energy reward shaping could reduce the energy expenditure in terms of a player's movements or actions. The lowest energy consuming action is the stop action (SPD: $0.0$), intermediate energy is required for maintaining the same speed and heading without accelerating or decelerating (HDG: $0.0$), while higher energies are required for steering or changing the heading of the player. A successful validation of the hypothesis would show that when energy reward shaping is used, the agent chooses actions with lower energy requirements more often than higher energy actions. In Figure~\ref{fig:action_heat_maps_dqn}, the left vertical axis shows the different actions available to the defender, the horizontal axis shows the different reward shaping functions with and without energy reward shaping, and the plot shows the frequency with which different actions are picked by the defender while using different reward shaping functions, in terms of a heat-map. We see that using only sparse rewards (SR) the player rarely uses the low energy stop or constant speed actions (dark square at SR-SPD: 0), but when energy reward shaping is added (SR+ERS), the stop action is chosen very often (bright square at (SR+ERS)-SPD: 0). Similarly, comparing boundary and boundary-tag reward shaping with and without energy reward shaping (SR+BRS vs. SR+BERS and SR+BTRS vs. SR+BTERS), we see that adding energy reward shaping the player is learning the energy conservation sub-task and picking the stop and constant speed actions more frequently over other, more energy consuming actions. These comparisons support our claim in Hypothesis 3 that additional objectives of a player like reducing energy consumption during playing can be controlled via a reward shaping function.

\begin{figure}[t]
     \begin{tabular}{cc}
     \vspace*{-0.1in}
    \hspace*{-0.3in} \includegraphics[width=2.3in]{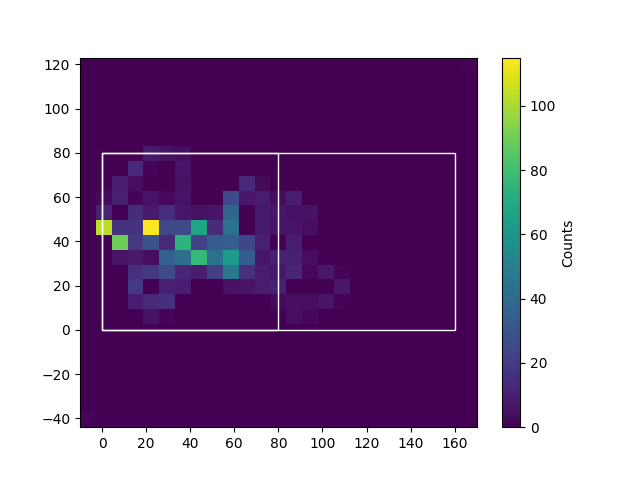} &
    \hspace*{-0.83in} \includegraphics[width=2.3in]{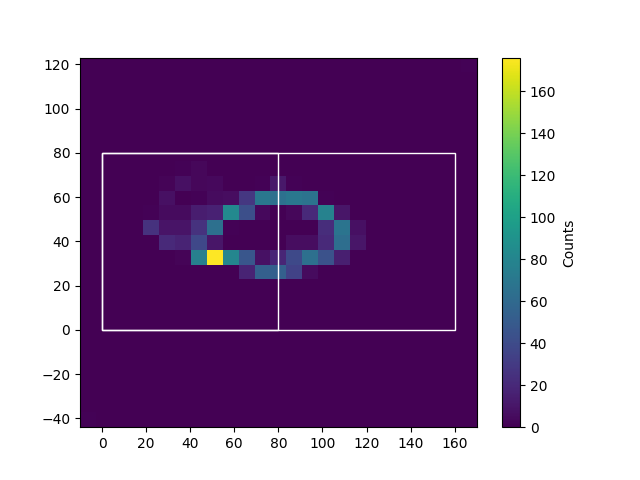}\\
    \hspace*{-0.6in}{\small SR + TRS ($10, 200$K)} & 
    \hspace*{-1.0in}{\small SR + $3$TRS ($26, 400$K)} \\   
     \vspace*{-0.1in}
    \hspace*{-0.35in}\includegraphics[width=2.3in]{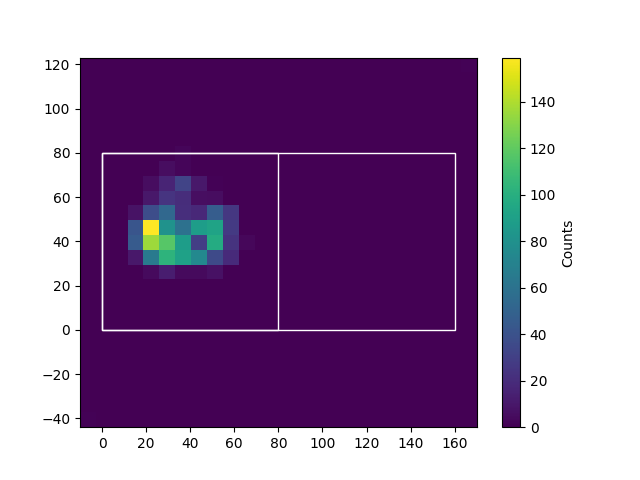} &
    \hspace*{-0.83in} \includegraphics[width=2.3in]{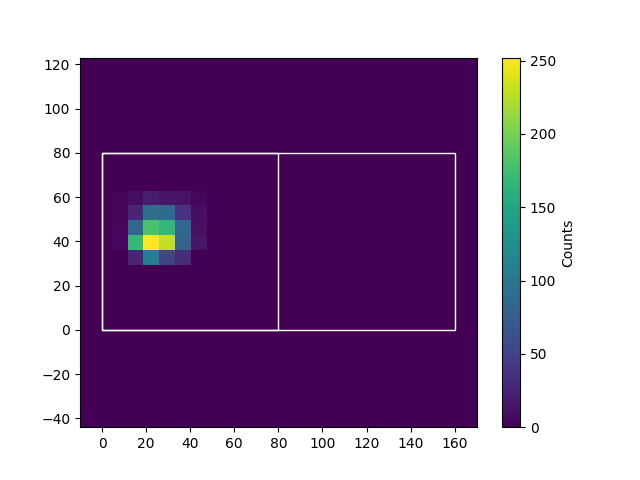} \\
    \hspace*{-0.6in}{\small SR + BRS ($18, 100$K)} & 
    \hspace*{-1.0in}{\small SR + $3$BRS ($24, 250$K)} \\

    \end{tabular}
    \caption{Position heat map showing locations in the playing field that are visited by defender playing against Att-E with tag-shaped reward with $1\times$ and $3\times$ gradient (top row) and boundary-shaped reward with $1\times$ and $3 \times$ gradient (bottom row).}
    \label{fig:pos_heat_map_bound_tag_3x}
\end{figure}

{\bf DQN Hypothesis 4.} Hypothesis 4's objective is to evaluate the affect of the reward shaping function's shape, determined by the function's gradient, on the reward convergence times and player performance. We changed the gradients of the tag and boundary reward shaping functions ($\kappa_2, \kappa_4$ in Equation~\ref{eqn:reward_shaping_bound} and $\omega_2, \omega_4$ in Equation~\ref{eqn:reward_shaping_tag}) to three times their original gradients. We expect $3\times$ gradient on the reward shaping functions to yield lower training times. For visualizing the reward shaping function gradient's effect, we plot position heat maps that show the frequency with which the defender visits different locations inside the playing field for different reward shaping functions, in Figure~\ref{fig:pos_heat_map_bound_tag_3x}. We see that when tag reward gradient goes from $1\times$ to $3\times$ (top row sub-plots in Figure~\ref{fig:pos_heat_map_bound_tag_3x}), the defender starts behaving more aggressively - it ignores the boundary sub-task and prioritizes the tag sub-task, resulting in it following the attacker's path (light blue trail in center of top right sub-plot), stopping just outside its zone border (bright yellow spot) to tag the attacker every time, as soon as it enters into the defender's zone. However, this behavior also effects more action changes, and, consequently, more energy expenditure by the defender, as evidenced by comparing the SR+TRS and SR+3TRS action heat maps in Figure~\ref{fig:action_heat_maps_dqn}. Changing the gradient of the boundary reward from $1\times$ to $3\times$ (bottom row sub-plots in Figure~\ref{fig:pos_heat_map_bound_tag_3x}) has the effect of contracting the area in which the defender moves around during playing. This is manifested by the lighter zone getting more concentrated near the defender's flag. The defender also chooses the stop action more frequently because the increased boundary gradient makes it learn that if it does the stop action within its base and near its flag more often, it can avoid entering or getting closer to areas near the boundary with negative rewards. Overall, we see that while the player's behavior can be modified significantly by changing the gradient of the reward shaping function, our Hypothesis 4, increasing reward function gradient ensures faster convergence times and higher rewards, is not fully supported. In some instances, the $3\times$ gradient resulted in a higher score than their non-increased counterpart, but the higher score came with significant increases in the training time. Increasing the gradient also resulted in poorer generality and performs worse against Att-H than with $1\times$ gradient. These findings point to the direction that reward shaping functions associated with different tasks interact with others in a complex, non-linear manner and need to be understood and designed carefully.

\begin{table*}[htb!]
\begin{center}
 \caption{\label{table:trained_e_vs_e}Performance of different reward shaping functions with defender trained against Att-E and evaluated against Att-E. First two columns and latter three columns of each sub-table show attacker's and defender's successful events resp.}
\begin{tabular}{|c|  c  c | c  c  c|| c  c | c  c  c||c  c | c  c  c||c  c | c  c  c||}
 \hline
& \multicolumn{2}{|c|}{{Att-E}} & \multicolumn{3}{|c||}{Def-E-{\tt SR}} & \multicolumn{2}{|c|}{{Att-E}} & \multicolumn{3}{|c||}{Def-E-{\tt SR+TRS}} & \multicolumn{2}{|c|}{{Att-E}} & \multicolumn{3}{|c||}{Def-E-{\tt SR+BRS}} & \multicolumn{2}{|c|}{{Att-E}} & \multicolumn{3}{|c||}{Def-E-{\tt SR+BTRS}}\\
 \hline
 Steps & Grb & Cap & Tag & Ret & Def & Grb & Cap & Tag & Ret & Def & Grb & Cap & Tag & Ret & Def & Grb & Cap & Tag & Ret & Def\\
     &    &      &     & Tag & Tag &     &     &     & Tag & Tag &     &      &    & Tag & Tag &    &      &     & Tag & Tag \\
 \hline
 \hline
 50k & 8 & 0 & 2 & 8 & 19 & 6 & 3 & 4 & 3 & 13 & 10 & 3 & 0 & 7 & 16 &\cellcolor{Yellow} 0 & \cellcolor{Yellow}0 & \cellcolor{Yellow}11 & \cellcolor{Yellow}0 & \cellcolor{Yellow}0\\
 100k & 9 & 9 & 1 & 0 & 8 & 3 & 3 & 7 & 0 & 14 & \cellcolor{Yellow}1 &\cellcolor{Yellow} 0 & \cellcolor{Yellow}9 & \cellcolor{Yellow}1 &\cellcolor{Yellow} 0  & 4 & 1 & 7 & 3 & 0 \\
 150k & 5 & 1 & 7 & 4 & 18 & 7 & 5 & 4 & 2 & 10 & 6 & 0 & 4 & 6 & 2 & 7 & 0 & 3 & 7 & 0  \\
 200k & 8 & 5 & 2 & 3 & 10 & \cellcolor{Yellow}2 &\cellcolor{Yellow} 2 & \cellcolor{Yellow}10 & \cellcolor{Yellow}0 &\cellcolor{Yellow} 4 & 2 & 0 & 8 & 2 & 2 & 6 & 1 & 4 & 5 & 2  \\
 250k & 9 & 7 & 1 & 1 & 12 & 9 & 7 & 1 & 0 & 16 & 5 & 0 & 5 & 5 & 0 & 4 & 1 & 7 & 2 & 5  \\
 300k & 10 & 8 & 0 & 2 & 16  & 3 & 2 & 9 & 0 & 9 & 8 & 6 & 2 & 2 & 2 & 9 & 4 & 1 & 5 & 6 \\
 350k & 9 & 8 & 1 & 1 & 12 & 3 & 3 & 7 & 0 & 9 & 6 & 0 & 4 & 6 & 0 & 6 & 1 & 4 & 5 & 0 \\
 400k & \cellcolor{Yellow}4 & \cellcolor{Yellow}0 & \cellcolor{Yellow}6 & \cellcolor{Yellow}4 &\cellcolor{Yellow} 0 & 7 & 5 & 3 & 2 & 10 & 10 & 8 & 0 & 1 & 11 & 5 & 2 & 5 & 3 & 8 \\
 \hline
\end{tabular}
\end{center}
\end{table*}

\begin{table*}
\begin{center}
\caption{Scores of Defender trained with different reward shaping functions against Att-E (top half) and Att-H (bottom half) and evaluated against Att-E (left sub-table) and Att-H (right sub-table)}
\label{table:trained_e_med_vs_e_med_score}
\hspace*{-0.1in}
\begin{tabular}{|c|c|c c c c||c c c c||}
 \hline
&  & \multicolumn{4}{c||}{vs. Att-E} & \multicolumn{4}{c||}{vs. Att-H} \\
  \hline
& Steps & {\tt SR} & {\tt TRS} & {\tt BRS} & {\tt BTRS} & {\tt SR} & {\tt TRS} & {\tt BRS} & {\tt BTRS}\\
 \hline
 \hline
{\multirow{8}{*}{{\rotatebox[origin=c]{90}{Def-E}}} }
& 50k & -15 & -14 & -7 & \cellcolor{Yellow}{\bf 22} & -31 & -29 & -55 & -20 \\
& 100k & -33 & -9 & \cellcolor{Yellow}{\bf 18} & 11 & -30 & -28 & -25 & -22 \\
& 150k & -7 & -17 & 6 & 6 & \cellcolor{Yellow}{\bf -21} & -17 & -28 & -27\\
& 200k & -21 & \cellcolor{Yellow}{\bf 10} & 14 & 3 & -25 & -22  & -12 & \cellcolor{Yellow}{\bf 4}\\
& 250k & -32 & -37  & 10& 5 & -32 & -16  & -20& -17 \\
& 300k & -40 & 2  & -16& -16 & -47 & -14 &\cellcolor{Yellow} {\bf -10} & -11 \\
& 350k & -34 & -4 & 8 & 5 & -41 & \cellcolor{Yellow}{\bf -1} & -18 & -24  \\
& 400k &\cellcolor{Yellow} {\bf 12} & -19  & -36& -4 & -23 & -7 & -38 & -24 \\
 \hline
  \hline
& Steps & {\tt SR} & {\tt TRS} & {\tt BRS} & {\tt BTRS} & {\tt SR} & {\tt TRS} & {\tt BRS} & {\tt BTRS}\\ 
\hline
 \hline
{\multirow{8}{*}{{\rotatebox[origin=c]{90}{Def-H}}} }
& 50k & \cellcolor{Yellow}{\bf -5} & -18  & 6 & -19 & -32 & -14 & 0 & -30\\
& 100k & -15 & -3 & 13 & -11 & -34 & -5 & \cellcolor{Yellow}{\bf 20} & \cellcolor{Yellow}{\bf -3}\\
& 150k & -45 & 7  & 20& \cellcolor{Yellow}{\bf 22} & -47 &  \cellcolor{Yellow}{\bf 1} & -11 & -8\\
& 200k & -21 & 1 & \cellcolor{Yellow}{\bf 22} & -3 & \cellcolor{Yellow}{\bf -30} & -1 & -27 & -10\\
& 250k & -25 & -11 & -14 & -3 & -36 & -6 & -3 & -5\\
& 300k & -46 & \cellcolor{Yellow}{\bf 18} & 0 & 14 & -42 & -11 & -20 & -14\\
& 350k & -13 & -49 & 3 & 16 & -29 & -16 & -22 & -8\\
& 400k & -30 & 3  & -20& -26 & -36 & -17 & -4 & -30\\
 \hline
\end{tabular}

\end{center}
\end{table*}

\end{document}